\PassOptionsToPackage{table}{xcolor}
\documentclass[twocolumn, 10pt, journal]{IEEEtran} 

\usepackage{cite}
\usepackage{amsmath}
\usepackage{amssymb}
\usepackage{amsfonts}
\usepackage{algorithmic}
\usepackage{graphicx}
\usepackage{textcomp}
\usepackage{subcaption}
\usepackage{booktabs}
\usepackage{makecell}
\usepackage{threeparttable}
\usepackage[shortcuts,acronym]{glossaries}
\usepackage{nicematrix}
\usepackage{multirow}
\usepackage{algorithm}
\usepackage{algorithmic}
\makeatletter
\newcommand\fs@norules{\def\@fs@cfont{\bfseries}\let\@fs@capt\floatc@ruled
  \def\@fs@pre{}%
  \def\@fs@post{}%
  \def\@fs@mid{\kern3pt}%
  \let\@fs@iftopcapt\iftrue}
\makeatother
\floatstyle{norules}
\restylefloat{algorithm}
\newacronym{auc}{AUC}{Area Under the Curve}
\newacronym{svd}{SVD}{Singular Value Decomposition}
\newacronym{mle}{MLE}{Maximum Likelihood Estimation}
\newacronym{ecg}{ECG}{Electrocardiography}
\newacronym{hr}{HR}{heart rate}
\newacronym{hrv}{HRV}{heart rate variability}
\newacronym{sqi}{SQI}{Signal Quality Index}
\newacronym{sqis}{SQIs}{Signal Quality Indices}
\newacronym{ppg}{PPG}{Photoplethysmography}
\newacronym{rmssd}{RMSSD}{Root Mean Square of Successive Differences}
\newacronym{sdnn}{SDNN}{Standard Deviation of RR-Intervals}
\newacronym{sdsd}{SDSD}{Standard Deviation of the Successive Differences}
\newacronym{rmsd}{RMSD}{Root Mean Square Deviation}
\newacronym{hf}{HF}{High Frequency}
\newacronym{lf}{LF}{Low Frequency}
\newacronym{pvcs}{PVCs}{premature ventricular contractions}
\newacronym{pvc}{PVC}{premature ventricular contraction}
\newacronym{rmse}{RMSE}{Root Mean Squared Error}
\newacronym{la}{LA}{Left Arm}
\newacronym{ra}{RA}{Right Arm}
\newacronym{ll}{LL}{Left Leg}
\newacronym{rl}{RL}{Right Leg}
\newcolumntype{C}{>{\centering\arraybackslash}X}
\newcolumntype{R}{>{\raggedleft\arraybackslash}X}
\newcolumntype{L}{>{\raggedright\arraybackslash}X}
\newacronym{ansi_aami}{ANSI/AAMI}{American National Standards Institute/Advancement of Medical Instrumentation}
\def\BibTeX{{\rm B\kern-.05em{\sc i\kern-.025em b}\kern-.08em
    T\kern-.1667em\lower.7ex\hbox{E}\kern-.125emX}}
\begin{document}
\title{Comprehensive Signal Quality Evaluation of a Wearable Textile ECG Garment: A~Sex-Balanced Study}
\author{Maximilian P. Oppelt, Tobias S. Zech, Sarah H. Lorenz, Laurenz Ottmann, Jan Steffan, \\ %
Bjoern M. Eskofier, Nadine R. Lang-Richter and Norman Pfeiffer %
\thanks{%
\textbf{This is a preprint of a manuscript submitted for publication. It has not yet been peer-reviewed, %
and the final version may differ.} \\
The authors acknowledge the funding by the EU TEF-Health project which is part %
of the Digital Europe Program of the EU (DIGITAL-2022-CLOUD-AI-02-TEFHEALTH). %
}%
\thanks{M. P. Oppelt (Main Contributing Author) is Senior Scientist at the Department Digital Health and Analytics, %
Fraunhofer IIS, Fraunhofer Institute for Integrated Circuits IIS, 91058 Erlangen, Germany and with the Department %
Artificial Intelligence in Biomedical Engineering, Friedrich-Alexander-University Erlangen Nuremberg, 91052 Erlangen, %
Germany %
(\emph{Corresponding e-mail: maximilian.oppelt@iis.fraunhofer.de}) %
} %
\thanks{Bjoern M. Eskofier is Professor at the Department Artificial Intelligence in Biomedical Engineering, %
Friedrich-Alexander-University Erlangen Nuremberg, 91052 Erlangen and Principal Investigator for Translational %
Digital Health Group at Institute of AI for Health Helmholtz Zentrum München, 85764 Munich, Germany %
} %
\thanks{T. Zech, S.H. Lorenz, L. Ottmann, J. Steffan, N.R. Lang-Richter and N. Pfeiffer %
are with the Department Digital Health and Analytics, Fraunhofer IIS, Fraunhofer Institute for Integrated Circuits %
IIS, 91058 Erlangen, Germany } %
} %
\maketitle
\begin{abstract} %
We introduce a novel wearable textile-garment featuring an innovative electrode placement aimed at minimizing noise
and motion artifacts, thereby enhancing signal fidelity in \gls{ecg} recordings. We present a comprehensive,
sex-balanced evaluation involving 15 healthy males and 15 healthy female participants to ensure the device's
suitability across anatomical and physiological variations. The assessment framework encompasses distinct
evaluation approaches: quantitative signal quality indices to objectively benchmark device performance; rhythm-based
analyzes of physiological parameters such as \gls{hr} and \gls{hrv}; machine learning classification tasks to assess
application-relevant predictive utility; morphological analysis of \gls{ecg} features including amplitude and interval
parameters; and investigations of the effects of electrode projection angle given by the textile / body shape, with all
analyzes stratified by sex to elucidate sex-specific influences. Evaluations were conducted across various activity
phases representing real-world conditions. The results demonstrate that the textile system achieves signal quality
highly concordant with reference devices in both rhythm and morphological analyses, exhibits robust classification
performance, and enables identification of key sex-specific determinants affecting signal acquisition. These findings
underscore the practical viability of textile-based \gls{ecg} garments for physiological monitoring as well as 
psychophysiological state detection. Moreover, we identify the importance of incorporating sex-specific design
considerations to ensure equitable and reliable cardiac diagnostics in wearable health technologies.
\end{abstract}
\begin{IEEEkeywords}
Electrocardiography, Garments, Machine Learning, Signal Quality, Smart Textiles, Wearables
\end{IEEEkeywords}
\section{Introduction}
\label{sec:introduction}
\IEEEPARstart{E}{lectrocardiographic} recordings serve as a fundamental diagnostic tool in modern medicine, providing
invaluable noninvasive insights into the electrical activity of the heart and therefore the health of the cardiovascular
system. Introduced by Willem Einthoven in the early 20th century, \acrfull{ecg} remains a cornerstone in clinical
cardiology. Einthoven's pioneering work laid the foundation for understanding the principles underlying ECG acquisition
and interpretation \cite{einthovenGalvanometrischeRegistrirungMenschlichen1903,%
kligfieldRecommendationsStandardizationInterpretation2007}.

\gls{ecg} signals are acquired through electrodes placed on the skin, capturing the electrical impulses generated by
cardiac muscle de- and repolarization. These signals manifest as characteristic waves on the \gls{ecg} tracing,
reflecting the sequential activation of different regions of the heart
\cite{reisnerPhysiologicalBasisElectrocardiogram2006}.

In modern medicine, \gls{ecg} is used in applications ranging from diagnosing cardiac arrhythmias
\cite{masonRecommendationsStandardizationInterpretation2007} and ischemic heart disease
\cite{sternDiagnosticAccuracyAmbulatory1975} to monitoring patients during surgery 
\cite{eichhornStandardsPatientMonitoring1986} and assessing the effects of pharmacological interventions
\cite{malikEvaluationDrugInducedQT2001, demmelNoQTcProlongation2018}. Moreover, recent advancements in wearable devices
have expanded ECG applications beyond traditional cardiology, enabling emerging research in affective sensing
\cite{oppeltADABaseMultimodalDataset2022} and stress detection during daily activities
\cite{sahaInvestigationRelationPhysiological2021}.

% Current state of electrocardiographic recordings
Each of these applications have domain specific requirements regarding the quality of the \gls{ecg} signal,
the duration of the recording, usability, accessibility, the comfort of the patient and costs.
While the traditional approach of using Ag/AgCl gel electrodes provides reliable signal quality, it also presents
several limitations.
These include the risk of skin irritation \cite{hanselContactSensitivityElectrocardiogram2020,
fotiContactAllergyElectrocardiogram2018}, the limitation that only trained professionals are able to correctly
apply the adhesive electrodes at specific anatomical locations \cite{crawfordPracticalAspectsECG2012}, and restricted
freedom of movement imposed on patients during daily activities by the presence of electrodes and connecting wires.

% General paragraph about wearable technology
Recent advances in wearable technology have addressed specific challenges in health monitoring applications. Devices
such as smartwatches \cite{evensonReviewValidityReliability2020} and smart rings \cite{caoAccuracyAssessmentOura2022}
utilize \gls{ppg} to measure \acrfull{hr} and \acrfull{hrv}, thereby enabling applications such as activity tracking
\cite{el-amrawyAreCurrentlyAvailable2015} and stress monitoring \cite{peakeCriticalReviewConsumer2018}. Despite their
utility, these devices have limited accuracy in capturing the precise timing of heartbeats, as they measure the
mechanical pulse of blood flow rather than the heart's electrical activity. Consequently, their diagnostic capabilities
are restricted, particularly for conditions such as ischemic heart disease or conduction disorders, as well as for
obtaining accurate \gls{hrv} parameter estimates, all of which require precise beat segmentation. Furthermore, their
measurements are susceptible to noise arising from individual patient characteristics such as skin tone, weight, age,
and gender as well as physiological factors including blood pressure and temperature, and external influences like
ambient lighting conditions \cite{fineSourcesInaccuracyPhotoplethysmography2021}.
Another category of devices, such as the Apple Watch \cite{aneeshs.FDAElectrocardiographSoftware2020}, is capable of
recording brief segments of \gls{ecg} waveforms and can automatically detect arrhythmias
\cite{haverkampVorhofflimmerndiagnostikMittelsEKGfaehiger2022, veltmannWearablebasierteDetektionArrhythmien2021}.
Similarly, systems like AliveCor \cite{bhargavaAliveCor2018, halcoxAssessmentRemoteHeart2017} are utilized to assess
morphological changes during pharmacological studies \cite{chungQTCIntervalsCan2015}.
Smart textiles represent a unique class of wearable devices that can continuously record the heart's electrical
activity over extended periods \cite{nigusseWearableSmartTextiles2021}. Unlike conventional wearables, smart textiles
offer enhanced comfort and freedom of movement, enabling unobtrusive and continuous \gls{ecg} monitoring during daily
activities. This is achieved by seamlessly integrating flexible dry electrodes into the garment itself, thereby further
enhancing wearer comfort \cite{alizadeh-meghraziEvaluationDryTextile2021,
eskandarianRobustMultifunctionalConductive2020, acarWearableFlexibleTextile2019}. Various research groups and device
manufacturers have developed different materials for this purpose, like metal based contact surfaces
\cite{catrysseIntegrationTextileSensors2004, marquezComparisonDrytextileElectrodes2010, yooWearableECGAcquisition2009,
nigusseWearableSmartTextiles2021, tseghaiIntegrationConductiveMaterials2020, paniSurveyTextileElectrode2018},
conductive polymers \cite{paniFullyTextilePEDOT2016, chlaihawiDevelopmentPrintedFlexible2018,
nigusseWearableSmartTextiles2021, tseghaiIntegrationConductiveMaterials2020, paniSurveyTextileElectrode2018,
tsukadaValidationWearableTextile2019} and carbon materials \cite{leePUNanowebbasedTextile2019,
chlaihawiDevelopmentPrintedFlexible2018, tasneemLowPowerOnChipECG2020, nigusseWearableSmartTextiles2021}.
\begin{table*}[htb!]
    \centering
    \caption{Related work on textile-based ECG monitoring with the number of subjects, electrode setup, protocol, reference system, and selected quality assessment, sorted by date of publication. (F: Female, M: Male, U: Unknown)}
    \label{tab:related_work_textile}
    \renewcommand{\arraystretch}{1}
        \begin{threeparttable}
        \begin{tabular}{llllll}
        \toprule
        Ref.                                                  & \renewcommand{\arraystretch}{1}\begin{tabular}[c]{@{}l@{}}Study\\ Population\end{tabular}                      & \renewcommand{\arraystretch}{1}\begin{tabular}[c]{@{}l@{}}Electrodes (el)\\ Channels (ch.)\end{tabular}                           & Protocol                                                                                                                                                 & \renewcommand{\arraystretch}{1}\begin{tabular}[c]{@{}l@{}}Reference\\ System\end{tabular}                                   & \renewcommand{\arraystretch}{1}\begin{tabular}[c]{@{}l@{}}Quality\\ Assessment\end{tabular}                                                                     \\
        \midrule
        \cite{choPerformanceEvaluationTextileBased2011}       & \renewcommand{\arraystretch}{1}\begin{tabular}[c]{@{}l@{}}5 (5M),\\ healthy\end{tabular}                      & \renewcommand{\arraystretch}{1}\begin{tabular}[c]{@{}l@{}}3 el. (RA, LA, LL),\\ 2 ch. (mod. Einthoven)\end{tabular}               & \renewcommand{\arraystretch}{1}\begin{tabular}[c]{@{}l@{}}standing, sitting\\ (relaxed) \end{tabular}                                      & \renewcommand{\arraystretch}{1}\begin{tabular}[c]{@{}l@{}}Biopac MP150\\ (Ag/AgCl)\end{tabular}                             & \renewcommand{\arraystretch}{1}\begin{tabular}[c]{@{}l@{}}R-Peak detection rate\\ (estimated/observed)\end{tabular}                                             \\  
        \rowcolor[gray]{0.9} \cite{dirienzoEvaluationTextilebasedWearable2013}     & \renewcommand{\arraystretch}{1}\begin{tabular}[c]{@{}l@{}}40 (40M),\\ cardiovascular\\ diseases\end{tabular}  & 2 (V5, V5R)                                                                                                                          & \renewcommand{\arraystretch}{1}\begin{tabular}[c]{@{}l@{}}20 subjects (in bed),\\ 20 subjects (mild exercise,\\ cycloergometer, walking)\end{tabular} & \renewcommand{\arraystretch}{1}\begin{tabular}[c]{@{}l@{}}CL Delta / Fukuda\\ Denshi DS 5700 \\(Ag/AgCl)\end{tabular}       & \renewcommand{\arraystretch}{1}\begin{tabular}[c]{@{}l@{}}cardiologist rating (artifacts, \\ morphological, rhythm), \\ compute rule based rating\end{tabular}  \\
        \cite{leeFlexibleCapacitiveElectrodes2014}            & \renewcommand{\arraystretch}{1}\begin{tabular}[c]{@{}l@{}}4 (4M),\\ healthy\end{tabular}                      & \renewcommand{\arraystretch}{1}\begin{tabular}[c]{@{}l@{}}2 el. (chest belt),\\ 1 ch. (custom)\end{tabular}                       & \renewcommand{\arraystretch}{1}\begin{tabular}[c]{@{}l@{}}standing, treadmill\\ (4, 5, 6, 7 km/h)\end{tabular}                                        & \renewcommand{\arraystretch}{1}\begin{tabular}[c]{@{}l@{}}Biopac MP150\\ (Ag/AgCl)\end{tabular}                             & \renewcommand{\arraystretch}{1}\begin{tabular}[c]{@{}l@{}}QRS delination, sensitivity,\\ accuracy, TP, FP, FN\end{tabular}                                      \\
        \rowcolor[gray]{0.9} \cite{wederEmbroideredElectrodeSilver2015}            & \renewcommand{\arraystretch}{1}\begin{tabular}[c]{@{}l@{}}12* (12U),\\ unknown\end{tabular}                   & \renewcommand{\arraystretch}{1}\begin{tabular}[c]{@{}l@{}}3 el. (reference + 2 custom),\\ 1 ch. (custom)\end{tabular}             & rest and unknown motion                                                                                                                                  & \renewcommand{\arraystretch}{1}\begin{tabular}[c]{@{}l@{}}Synmedic Ambu\\ Blue (Ag/AgCl)\end{tabular}                       & subjective rating                                                                                                                \\
        \cite{paniFullyTextilePEDOT2016}                      & \renewcommand{\arraystretch}{1}\begin{tabular}[c]{@{}l@{}}10 (10U),\\ healthy\end{tabular}                    & \renewcommand{\arraystretch}{1}\begin{tabular}[c]{@{}l@{}}various\\ (task dependent)\end{tabular}                                 & \renewcommand{\arraystretch}{1}\begin{tabular}[c]{@{}l@{}}sitting, breathing, stairs,\\ 2Hz step-up-down-exercice\end{tabular}                        & -                                                                                                                              & QRS detection rate, RMSE \\ % \gls{rmse}                                                                                                          \\
        \rowcolor[gray]{0.9} \cite{boehmNovel12LeadECG2016}                        & \renewcommand{\arraystretch}{1}\begin{tabular}[c]{@{}l@{}}3 (3M),\\ unknown\end{tabular}                      & \renewcommand{\arraystretch}{1}\begin{tabular}[c]{@{}l@{}}9 el (chest),\\ 8 ch. (12 Mason-Likar)\end{tabular}                     & lying, sitting, walking                                                                                                                                  & \renewcommand{\arraystretch}{1}\begin{tabular}[c]{@{}l@{}}Corscience Holter\\ ECG BT12 (Ag/AgCl)\end{tabular}               & \renewcommand{\arraystretch}{1}\begin{tabular}[c]{@{}l@{}}Bland-Altmann, \\ fiducial points, \\ reference correlation\end{tabular}                              \\
        \cite{xiaoPerformanceEvaluationPlain2017}             & \renewcommand{\arraystretch}{1}\begin{tabular}[c]{@{}l@{}}10 (10U),\\ healthy\end{tabular}                    & \renewcommand{\arraystretch}{1}\begin{tabular}[c]{@{}l@{}}3 el (RA, RA, LL\tnote{$\dagger$} ),\\ 1 ch. (mod. Lead I)\end{tabular} & -                                                                                                                                                        & \renewcommand{\arraystretch}{1}\begin{tabular}[c]{@{}l@{}}Unknown\\ (Ag/AgCl)\end{tabular}                                  & \renewcommand{\arraystretch}{1}\begin{tabular}[c]{@{}l@{}}R-peak amplitude\\ and variations, SNR\end{tabular}                                                   \\
        \rowcolor[gray]{0.9} \cite{sunWearableHshirtExercise2017}                  & \renewcommand{\arraystretch}{1}\begin{tabular}[c]{@{}l@{}}4 (2M, 2F),\\ healthy, minors\end{tabular}         & \renewcommand{\arraystretch}{1}\begin{tabular}[c]{@{}l@{}}2 el. (chest),\\ 1 ch. (custom)\end{tabular}                            & \renewcommand{\arraystretch}{1}\begin{tabular}[c]{@{}l@{}}resting, walking,\\ running (12 km/h)\end{tabular}                                          & \renewcommand{\arraystretch}{1}\begin{tabular}[c]{@{}l@{}}Cardiac Monitor \\ Arm Band\\ (Ag/AgCl)\end{tabular}              & \renewcommand{\arraystretch}{1}\begin{tabular}[c]{@{}l@{}}visual inspection, HR, \\ tachycardia/extra-systoles\end{tabular}                                     \\
        \cite{chlaihawiDevelopmentPrintedFlexible2018}        & \renewcommand{\arraystretch}{1}\begin{tabular}[c]{@{}l@{}}1 (1U),\\ healthy\end{tabular}                      & \renewcommand{\arraystretch}{1}\begin{tabular}[c]{@{}l@{}}3 el (LA, RA, LL),\\ 1 ch. (mod. Lead I)\end{tabular}                   & -                                                                                                                                                        & \renewcommand{\arraystretch}{1}\begin{tabular}[c]{@{}l@{}}Analog Devices\\ INA 2128 (Ag/AgCl)\end{tabular}                  & \renewcommand{\arraystretch}{1}\begin{tabular}[c]{@{}l@{}}visual inspection, \\ reference signal correlation\end{tabular}                                       \\
        \rowcolor[gray]{0.9} \cite{anHybridTextileElectrode2018}                   & \renewcommand{\arraystretch}{1}\begin{tabular}[c]{@{}l@{}}1 (1F),\\ healthy\end{tabular}                      & \renewcommand{\arraystretch}{1}\begin{tabular}[c]{@{}l@{}}3 el (chest band),\\ 1 ch. (mod. Lead I)\end{tabular}                   & sitting, walking                                                                                                                                         & \renewcommand{\arraystretch}{1}\begin{tabular}[c]{@{}l@{}}ADS1292ECG-FE\\ (Ag/AgCl)\end{tabular}                            & \renewcommand{\arraystretch}{1}\begin{tabular}[c]{@{}l@{}}visual inspection,\\ baseline wander, HF noise\end{tabular}                                           \\
        \cite{achilliDesignCharacterizationScreenPrinted2018} & \renewcommand{\arraystretch}{1}\begin{tabular}[c]{@{}l@{}}4 (4U),\\ healthy\end{tabular}                      & \renewcommand{\arraystretch}{1}\begin{tabular}[c]{@{}l@{}}3 el (LA, RA, LL),\\ 1 ch. (mod. Lead I)\end{tabular}                   & -                                                                                                                                                        & \renewcommand{\arraystretch}{1}\begin{tabular}[c]{@{}l@{}}TMSI Proti7\\ (Ag/AgCl)\end{tabular}                              & \renewcommand{\arraystretch}{1}\begin{tabular}[c]{@{}l@{}}PSD, visual inspection, \\ delineation\end{tabular}                                                   \\
        \rowcolor[gray]{0.9} \cite{tsukadaValidationWearableTextile2019}           & \renewcommand{\arraystretch}{1}\begin{tabular}[c]{@{}l@{}}66 (47M, 19F\tnote{$*$} ), \\ healthy\end{tabular} & \renewcommand{\arraystretch}{1}\begin{tabular}[c]{@{}l@{}}2 el. (V5R, V5)\\ 1 ch. (custom location)\end{tabular}                  & \renewcommand{\arraystretch}{1}\begin{tabular}[c]{@{}l@{}}supine rest, seated rest,\\ trunk rotation, steps\end{tabular}                              & \renewcommand{\arraystretch}{1}\begin{tabular}[c]{@{}l@{}}Holter ECG Cardy 303\\ Pico+ (Ag/AgCl)\end{tabular}               & \renewcommand{\arraystretch}{1}\begin{tabular}[c]{@{}l@{}}R-amplitude variations\\ isoelectric line variations\end{tabular}                                     \\
        \cite{ankhiliAmbulatoryEvaluationECG2019}             & \renewcommand{\arraystretch}{1}\begin{tabular}[c]{@{}l@{}}1 (1F),\\ healthy\end{tabular}                      & \renewcommand{\arraystretch}{1}\begin{tabular}[c]{@{}l@{}}3 el (LA, RA, LL), \\ 1 ch. (Lead I)\end{tabular}                       & resting                                                                                                                                                  & -                                                                                                                              & visual inspection, SNR                                                                                                           \\
        \rowcolor[gray]{0.9} \cite{tasneemLowPowerOnChipECG2020}                   & \renewcommand{\arraystretch}{1}\begin{tabular}[c]{@{}l@{}}1 (1U),\\ healthy\end{tabular}                      & \renewcommand{\arraystretch}{1}\begin{tabular}[c]{@{}l@{}}3 el. (LA, RA, LL),\\ 1 ch. (mod. Lead I)\end{tabular}                  & -                                                                                                                                                        & -                                                                                                                              & \renewcommand{\arraystretch}{1}\begin{tabular}[c]{@{}l@{}}peak amplitude,\\ SNR (PSD)\end{tabular}                                                              \\
        \cite{liuIntegratedDesignMultiChannel2020}            & \renewcommand{\arraystretch}{1}\begin{tabular}[c]{@{}l@{}}1 (1U),\\ healthy\end{tabular}                      & \renewcommand{\arraystretch}{1}\begin{tabular}[c]{@{}l@{}}5 el. (LA, RA, LL, RL, V2),\\ 4 ch. (mod. Einthoven, V2)\end{tabular}   & \renewcommand{\arraystretch}{1}\begin{tabular}[c]{@{}l@{}}sitting, standing, walking,\\ stair climbing\end{tabular}                                   & \renewcommand{\arraystretch}{1}\begin{tabular}[c]{@{}l@{}}Texas Instruments\\ ADS1298R (Ag/AgCl)\end{tabular}               & visual inspection                                                                                                                \\
        \rowcolor[gray]{0.9} \cite{arquillaDetectionCompleteECG2021}               & \renewcommand{\arraystretch}{1}\begin{tabular}[c]{@{}l@{}}10 (10M),\\ healthy\end{tabular}                    & chest, unknown                                                                                     & relaxed, resting                                                                                                                                                                           & -                                                                                                                              & \renewcommand{\arraystretch}{1}\begin{tabular}[c]{@{}l@{}}PQRST delineation\\ (recall, precision, F1)\end{tabular}                                              \\
        \cite{alizadeh-meghraziEvaluationDryTextile2021}      & \renewcommand{\arraystretch}{1}\begin{tabular}[c]{@{}l@{}}1 (1U),\\ healthy\end{tabular}                      & \renewcommand{\arraystretch}{1}\begin{tabular}[c]{@{}l@{}}3 el. (LA, RA, LL),\\ 1 ch. (mod. Lead I)\end{tabular}                  & resting                                                                                                                                                  & -                                                                                                                              & \renewcommand{\arraystretch}{1}\begin{tabular}[c]{@{}l@{}}peak amplitude,\\ pSQI, BasSQI\end{tabular}                                                           \\
        \rowcolor[gray]{0.9} \cite{blasingECGPerformanceSimultaneous2022}          & \renewcommand{\arraystretch}{1}\begin{tabular}[c]{@{}l@{}}13 (6M, 7F),\\ healthy\end{tabular}                & \renewcommand{\arraystretch}{1}\begin{tabular}[c]{@{}l@{}}3 el. (LA, RA, LL, RL)\\ 2 ch. (mod. Einthoven)\end{tabular}            & \renewcommand{\arraystretch}{1}\begin{tabular}[c]{@{}l@{}}standing, treadmill,\\$n$-back task\end{tabular}                                            & \renewcommand{\arraystretch}{1}\begin{tabular}[c]{@{}l@{}}Bittium Faros 360\\NeXus-10 MKII\\Polar RS800 Multi\\SOMNOtouch NIBP\end{tabular}        & \renewcommand{\arraystretch}{1}\begin{tabular}[c]{@{}l@{}}morphological, \gls{hr} \end{tabular}                                    \\
        \cite{neriComparisonSingleLeadECG2024}                & \renewcommand{\arraystretch}{1}\begin{tabular}[c]{@{}l@{}}30 (16M, 14F),\\ healthy\end{tabular}              & \renewcommand{\arraystretch}{1}\begin{tabular}[c]{@{}l@{}}2 el., 1 ch. \end{tabular}                                              & \renewcommand{\arraystretch}{1}\begin{tabular}[c]{@{}l@{}}activity during daily \\ living (unspecified)\end{tabular}                                   & \renewcommand{\arraystretch}{1}\begin{tabular}[c]{@{}l@{}}3-lead holter \\monitor\end{tabular}                                                      & \renewcommand{\arraystretch}{1}\begin{tabular}[c]{@{}l@{}}subjective, R-R \\ interval comparison \end{tabular} \\
        \rowcolor[gray]{0.9} \textbf{Ours}                                         & \renewcommand{\arraystretch}{1}\begin{tabular}[c]{@{}l@{}}30 (15M, 15F),\\ healthy\end{tabular}                & \renewcommand{\arraystretch}{1}\begin{tabular}[c]{@{}l@{}}4 el. (LA, RA, LL, RL)\\ 2 ch. (mod. Einthoven)\end{tabular}            & \renewcommand{\arraystretch}{1}\begin{tabular}[c]{@{}l@{}}lying, sitting,\\ walking, running\end{tabular}                                             & \renewcommand{\arraystretch}{1}\begin{tabular}[c]{@{}l@{}}Bittium Faros 180 \&\\ BTL-08 Holter H600 \\ (Ag/AgCl)\end{tabular}                      &                                                                                                           \\
        \bottomrule
        \end{tabular}
        \begin{tablenotes}\footnotesize
            \item[$*$] Female subjects wore a different type of garment (bra instead of shirt) and electrode setup.
            \item[$\dagger$] Ag/AgCl electrode as a reference electrode.
            \end{tablenotes}
    \end{threeparttable}

\end{table*}

These conductive materials are integrated into the textile using various fabrication techniques: Embroideries, 
knittings, sewings and weavings are used to classically integrate yarns
\cite{polaTextileElectrodesECG2007, qinNovelWearableElectrodes2018, tseghaiIntegrationConductiveMaterials2020,
gunnarssonSeamlesslyIntegratedTextile2023}. Others have printed \cite{biharFullyPrintedElectrodes2017},
electrostatically flocked \cite{takeshitaRelationshipContactPressure2019}, or dip coated
\cite{leePUNanowebbasedTextile2019} conductive materials onto the textiles. Table~\ref{tab:related_work_textile}
summarizes several studies that have introduced novel materials and integration techniques for textile electrodes.
These efforts seek to optimize the signal-to-noise ratio, lower skin-electrode impedance, reduce motion artifacts
through robust and stable skin-electrode contact, increase durability, including machine washability and maintain
biocompatibility.

Significant differences exist in physiological parameters between male and female \gls{ecg} recordings. For instance,
women typically exhibit a higher \gls{hr}, a lower ST-segment, and a reduced T-wave amplitude compared to men
\cite{yarnozMoreReasonsWhy2008, kittnarSexRelatedDifferences2023}. Other anatomical factors,
including variations in chest shape affecting electrode placement and differences in heart size 
\cite{prajapatiSexDifferencesHeart2022}, can result in ambiguous recordings, potentially leading
to invalid diagnoses with certain electrode positions \cite{rautaharjuStandardizedProcedureLocating1998}.
When developing a new system intended for use by both sexes, it is crucial to evaluate the system with a diverse study
population. This ensures that the system is not biased and, as a result, can provide valid diagnostics
\cite{colacoFalsePositiveECG2000, drewPitfallsArtifactsElectrocardiography2006}.

A primary motivation for developing new wearable sensor technology is to enable the continuous acquisition of
high-quality \gls{ecg} recordings. However, current research in this field reveals several limitations, including small
sample sizes, study populations restricted to a single gender (typically male), and evaluation methodologies that focus
exclusively on electrode materials rather than assessing the signal quality of the entire wearable system.
These shortcomings, as detailed in Table~\ref{tab:related_work_textile}, give rise to two primary research questions:
\textbf{RQ1)} How does a wearable \gls{ecg} textile with integrated dry silicone electrodes perform in
accurately measuring ECG parameters like \gls{hr}, \gls{hrv}, and morphology compared to a medical-grade gold-standard
Holter ECG during different activities like sedentary tasks versus dynamic movements like walking or running?
\textbf{RQ2)} To what degree does the structural design of the textile-based ECG system influence
the quality of recorded signals, particularly in regard to addressing potential gender-related
disparities in cardiac diagnostics?
\section{Methods}
To evaluate the functionality and accuracy of our textile-based \gls{ecg} device, participants simultaneously wore our
wearable textile shirt with integrated dry electrodes, as well as two reference systems representing the current gold
standard, each employing Ag/AgCl electrodes.
\begin{figure}[ht!]
    \centering
    \begin{subfigure}[t]{0.325\linewidth}
        \includegraphics[width=\textwidth]{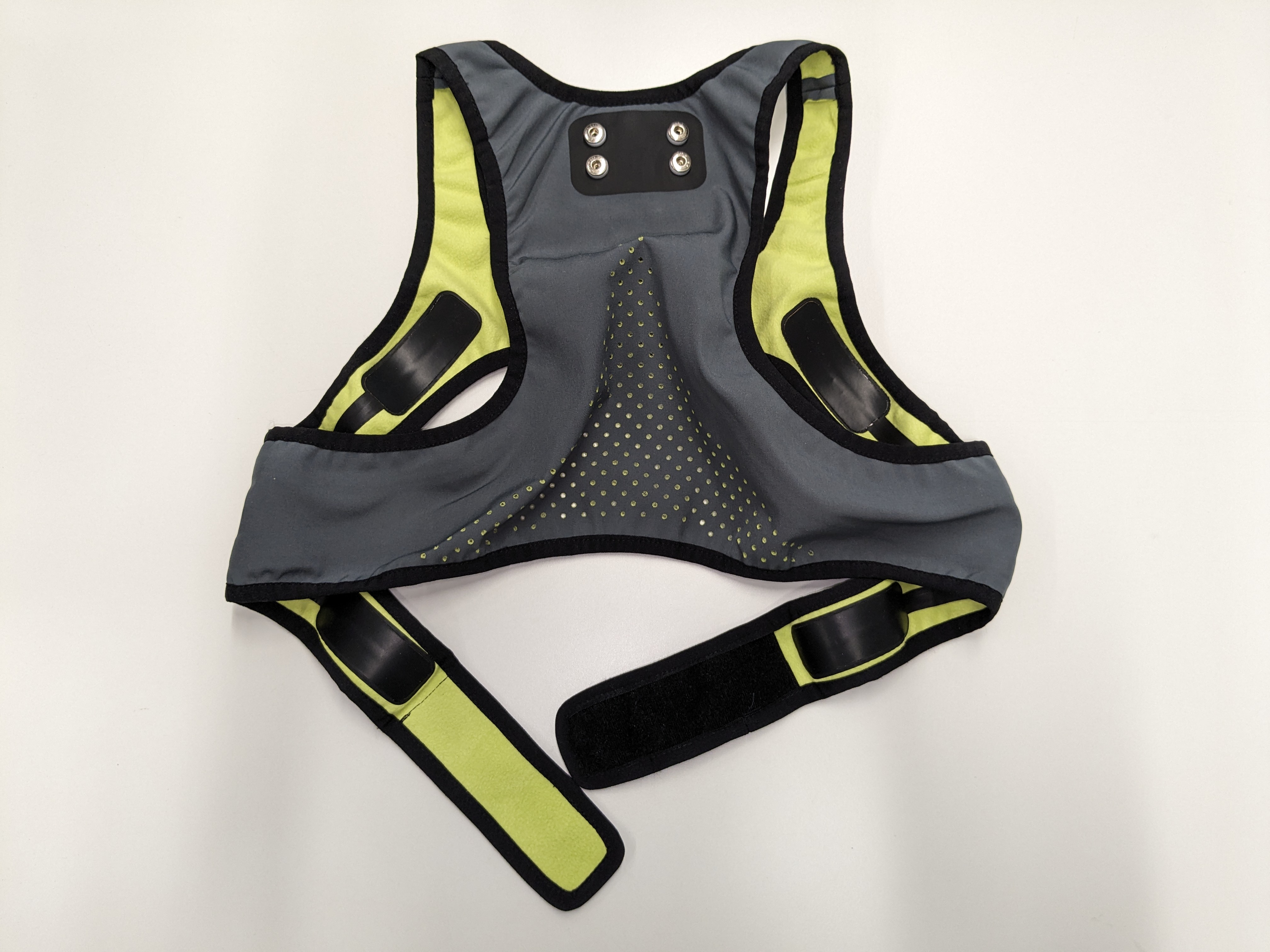}
    \end{subfigure}
    \begin{subfigure}[t]{0.325\linewidth}
        \includegraphics[width=\textwidth]{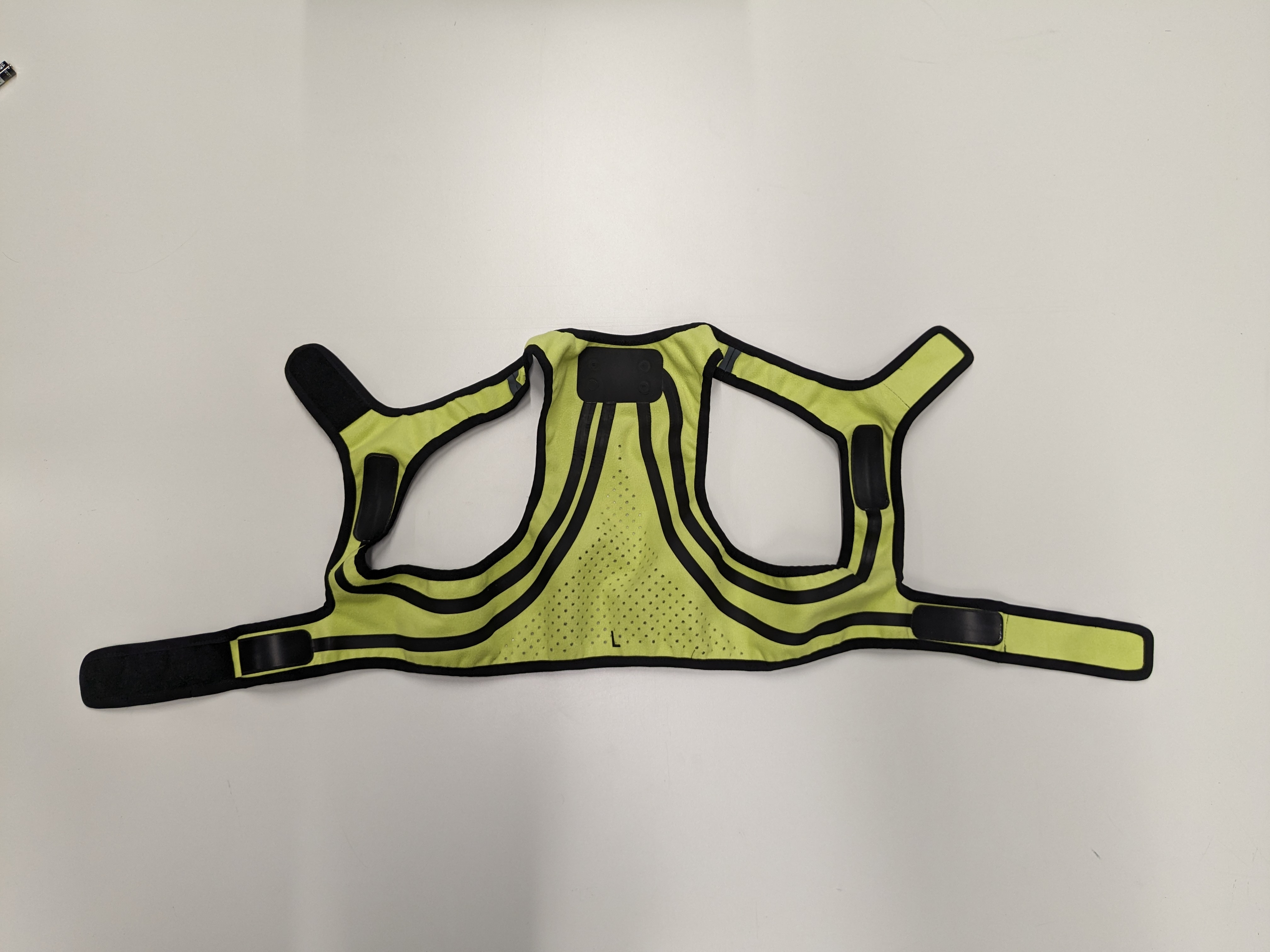}
    \end{subfigure}
    \begin{subfigure}[t]{0.325\linewidth}
        \includegraphics[width=\textwidth]{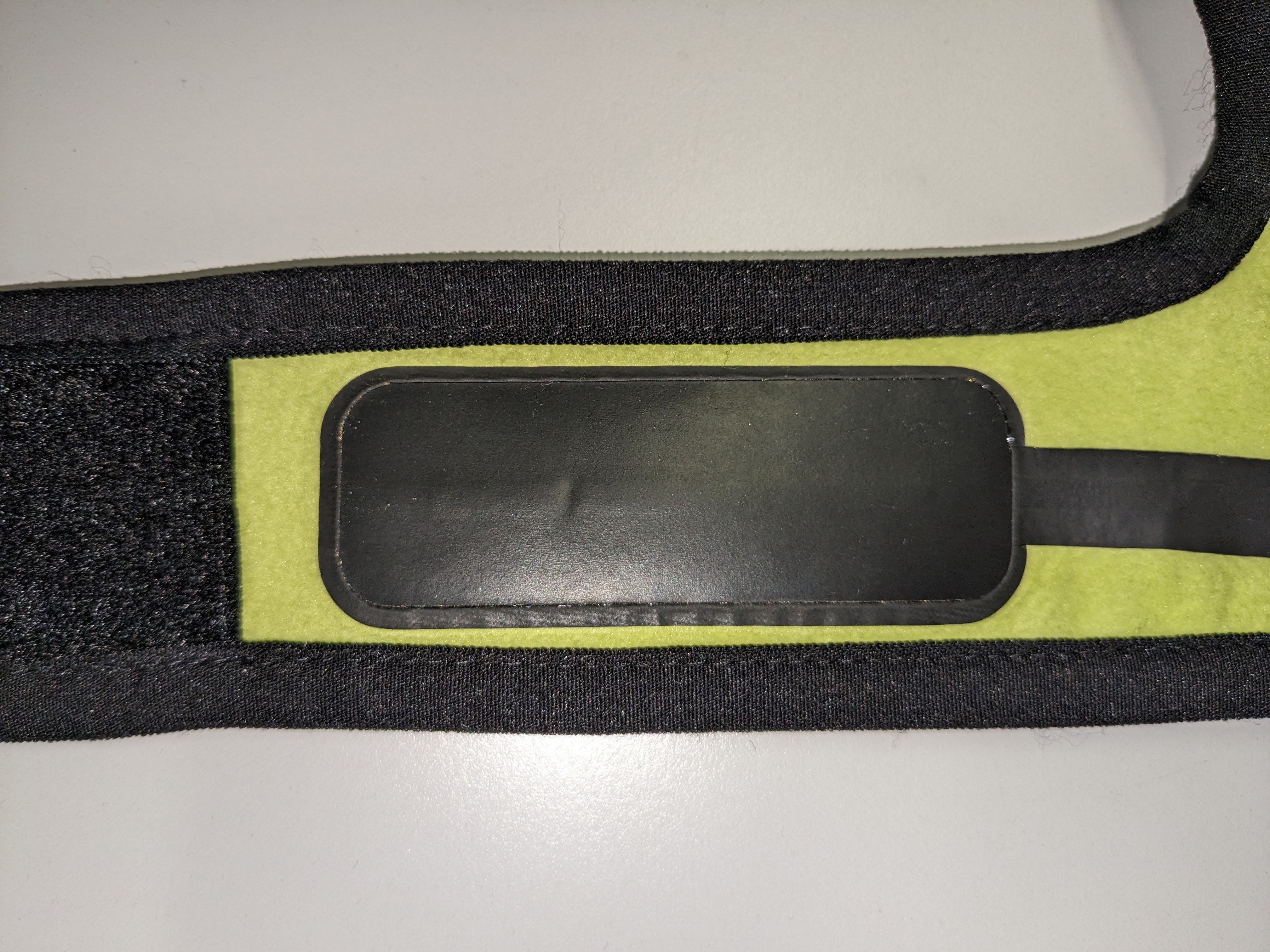}
    \end{subfigure}
    \caption{\small Wearable textile with integrated electrodes and connectors: rear view with button connectors
(left), interior wiring (middle), and close-up of flexible dry electrode (right).}
    \label{fig:garment-views}
    \vspace{-0.1cm}
\end{figure}
The textile incorporates four electrodes \gls{la}, \gls{ra}, \gls{ll}, and \gls{rl} providing three bipolar
leads (two directly measured and one calculated), corresponding to modified Einthoven leads recorded at the torso.
Figure~\ref{fig:garment-views} provides an overview of the garment with integrated electrodes and connections. The
fourth electrode serves as the neutral electrode. Flexible electrodes entail a multilayer structure comprising $78\%$
polyamide and $22\%$ elastomer stretch-tricot, metal-plated with silver ($26.9 \pm 1\%$), topped off by a conductive
silicone sheet. This composition ensures durability and low surface resistance, while its stretch properties provide
comfort during wear.
The electrode cables are seamlessly integrated into the textile and routed to a central hub positioned at the
back of the subject’s neck, where the connectors are located. These connectors feature buttons to which the hardware
can be attached. The electronics are responsible for digitizing the signal using a ADS1293 (Texas Instruments, USA)
analog front-end. The data storage on an SD card is handled by the microcontroller nRF52840 (Nordic Semiconductor,
Norway). During setup, the live signal is monitored via Bluetooth transmission, while recorded data stored on the SD
card is used for subsequent evaluation, ensuring data consistency without potential loss of wireless connections.
% Electorde material
% https://www.shieldex.de/products/shieldex-silitex/
% Shieldex Silitex
% 78% Polyamide + 22% Elastane
% Strech-Tricot
% Metal Plated with Silver of 26.9 \pm 1
% coated in conductive silicon
% with a electrical surface resistivity of < 5 Ohm/
% warp (100 %) & weft (150 %) double strech direction (DIN 13934-1)
% total weight of 380 g/m^2
As reference systems, we used two commercially available medical devices: the BTL (BTL GmbH, Dornstadt,
Germany) BT-08 Holter monitor and the Bittium (Bittium Corp., Oulu, Finland) Faros 180. Both devices are approved
for medical use and were operated with Ag/AgCl wet electrodes.
\subsection{Study Population}
The study population consists of $30$ subjects, with $15$ males and $15$ females. The minimal age is $21$, the maximal age
is $60$, and the mean age is $30.2$ with a standard deviation of $8.4$. The height distribution for males and
females, as well as the weight distribution, is $181.7 \pm 5.1\,\mathrm{cm}$, $168.2 \pm 6.8\,\mathrm{cm}$ and
$84.7 \pm 11.3\,\mathrm{kg}$, $60.7 \pm 7.4\,\mathrm{kg}$ respectively. We measured the under chest girth according to
the ISO 8559 \cite{DINISO85591} standard. The lower chest girth is $93.9 \pm 7.5\,\mathrm{cm}$ and
$75.9 \pm 4.6\,\mathrm{cm}$ for males and females respectively.
All subjects participated voluntarily, had no known cardiovascular disease and were not taking any medication,
except oral contraceptives, that could influence the \gls{ecg} signal. The study was conducted in accordance with the
Declaration of Helsinki and was approved by the Ethics Committee of Friedrich-Alexander-University Erlangen-Nuremberg
(\textit{65}\_\:\textit{21 B}, approved on \textit{16.03.2021}).
\subsection{Measurement Locations and Reference Systems}
\begin{figure}[ht!]
    \centering
    \begin{subfigure}[t]{0.49\linewidth}
        \includegraphics[width=\textwidth]{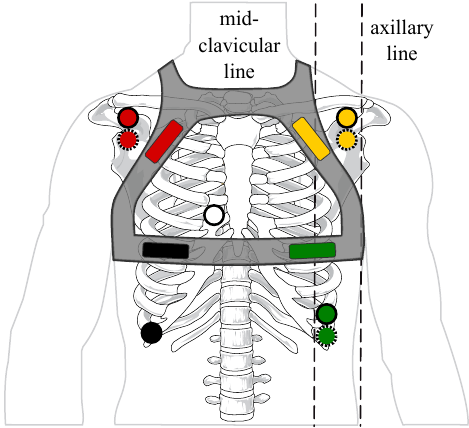}
        \caption{\small Reference system electrodes positioned at modified Einthoven locations 
(LA, RA, LL, RL) and the textile electrode location, with the shaded garment area.}
        \label{sfig:electrode-locations}
    \end{subfigure}
    \begin{subfigure}[t]{0.49\linewidth}
        \includegraphics[width=\textwidth]{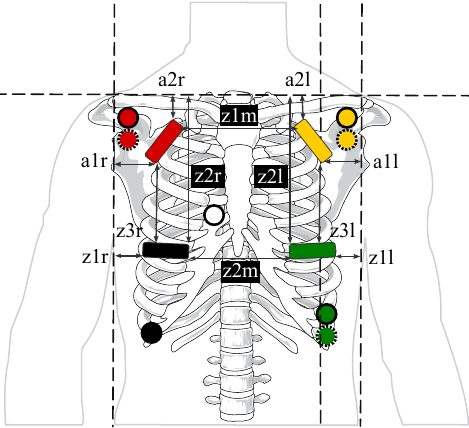}
        \caption{\small Locations of the reference electrodes and the textile, illustrating the spatial
relationship between the textile electrodes and the reference electrodes.}
        \label{sfig:electrode-measurements}
    \end{subfigure}
\caption{Electrode locations, fit and garment overview.}
    \vspace{-0.1cm}
\end{figure}

Since Einthoven introduced the three standard limb leads, several modifications have been made to improve the
practicality of lead placement. For example, during treadmill \gls{ecg} exercise, electrodes are attached to the
back to reduce movement artifacts. In mobile Holter systems, the electrodes are usually placed on the torso in the
direction of the standard Einthoven limb leads to allow for greater freedom of movement.
In our reference systems, the upper electrodes were mounted along the anterior axillary line (over the bone) at the
level of the clavicle, while the lower electrodes were placed outside the anterior axillary line at the end of the
ninth costal cartilage. For the BTL five-lead system, the precordial electrode was positioned at the fourth intercostal
space on the right margin of the sternum (lead~V1). This electrode placement was selected because it is a common scheme
for Holter monitors and provides sufficient space to accommodate the textile system.
\begin{figure}[ht!]
    \centering
    \begin{subfigure}[t]{0.49\linewidth}
        \includegraphics[width=\textwidth]{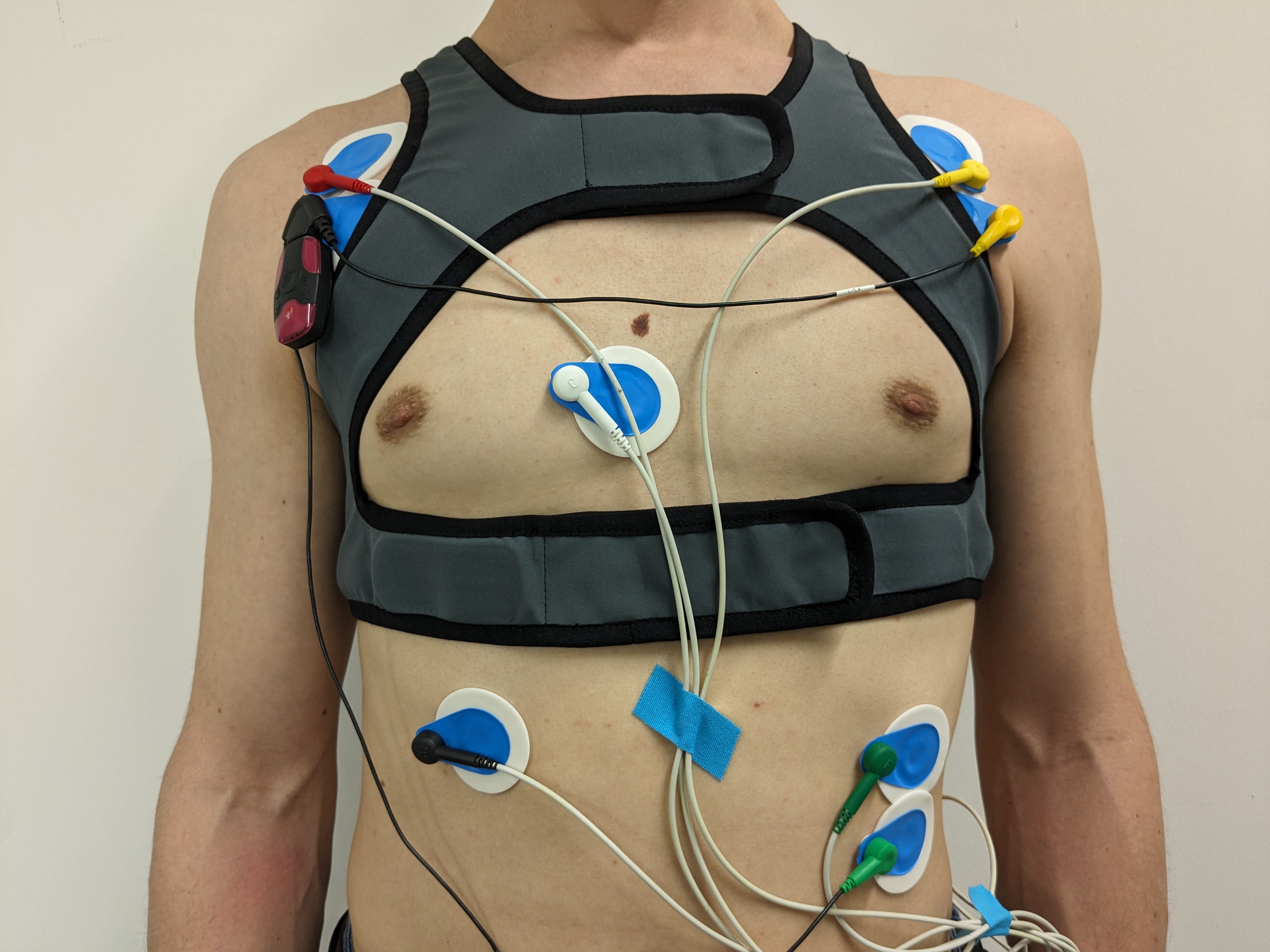}
        \caption{\small Subject wearing the garment with integrated and reference electrodes.}
        \label{sfig:example-mounts}
    \end{subfigure}
    \begin{subfigure}[t]{0.49\linewidth}
        \includegraphics[width=\textwidth]{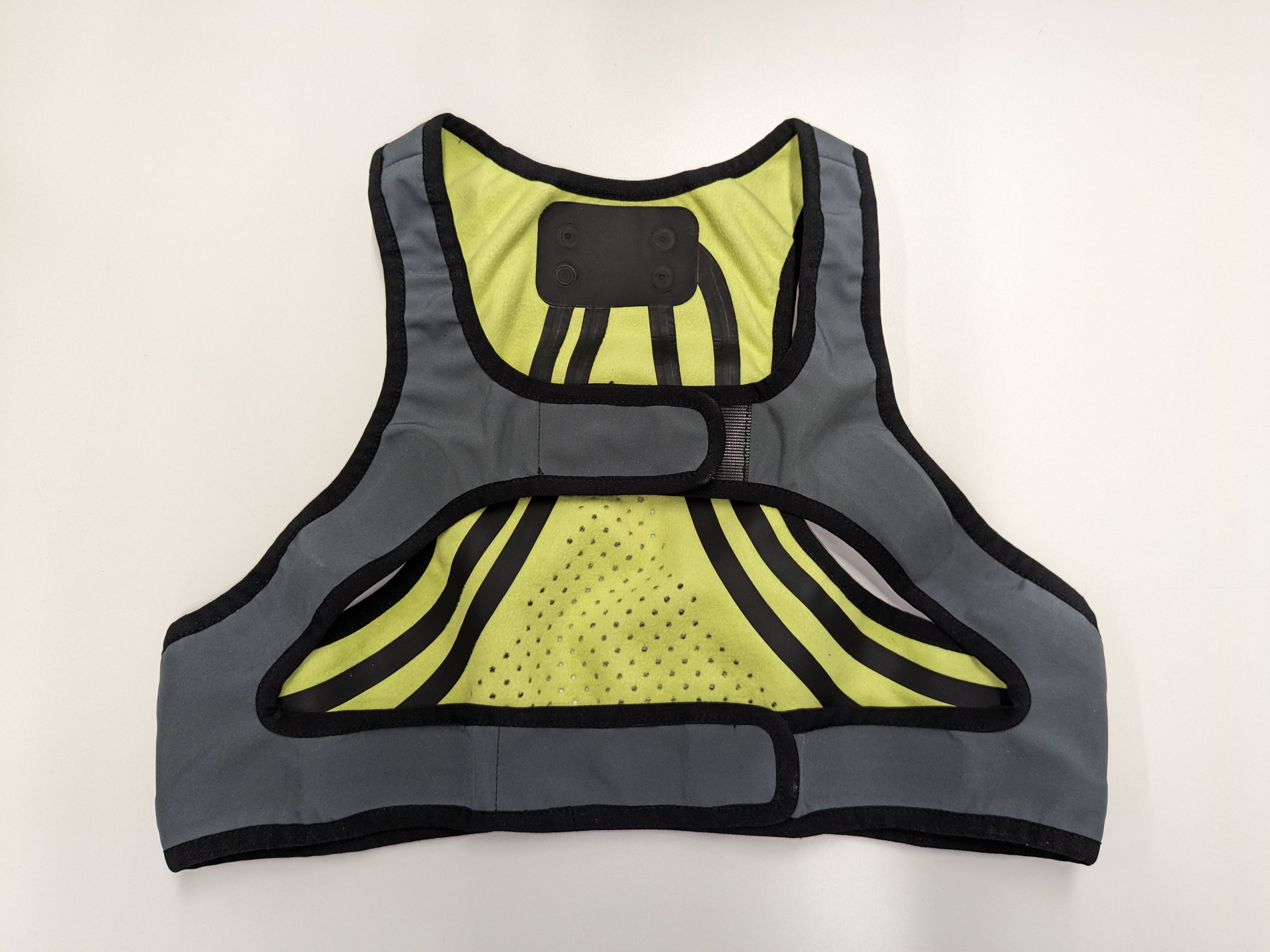}
        \caption{\small System overview of the wearable with adjustable hook-and-loop fastener.}
        \label{sfig:textile-overview}
    \end{subfigure}
    \caption{Male subject wearing garment, showing reference electrode placement from both systems.}
    \label{fig:positions}
    \vspace{-0.1cm}
\end{figure}
The textile electrodes are positioned closer to the heart in alignment with the limb leads. The lower electrodes
are placed at the fifth intercostal space on the anterior axillary line. These modifications are made to ensure
a tight fit and minimize movement artifacts resulting from limb motion due to the greater distance.
The garment was designed to ensure that the orientation of the electrodes closely aligns with natural physiological
contours. Specifically, the upper electrodes were positioned at a 45-degree angle relative to the axillary line,
while the lower electrodes were placed horizontally with a slight inward rotation shift towards the center.
This configuration allows the electrodes to fit to the torso without protruding, thereby following the body's
shape for optimal contact and comfort. 
The garment is designed with adjustable hook-and-loop fasteners positioned at both the upper and lower regions,
enabling precise adaptation to various body shapes and ensuring a tight fit. To accommodate more substantial variations
in body size, multiple garment sizes were produced in accordance with the DIN EN 13402-3:2017-12 standard, ranging
from XS to XXL.
The actual fit of the garment and the positioning of the hook-and-loop fasteners varies according to individual body
shape, which may, in turn, affect the placement of the electrodes.
To systematically assess electrode positioning, we measured several key distances, as illustrated in
Figure~\ref{sfig:electrode-locations} and Figure~\ref{sfig:electrode-measurements}. These figures depict the
electrode locations for both the reference and textile systems, as well as the measured distances between the
wearable electrodes and relevant anatomical landmarks. This approach enabled us to evaluate electrode positioning and
analyze the impact on the anatomical projections.
\subsection{Study Protocol}
The study protocol comprised a series of four-minute tasks, beginning with an initial seated resting phase, followed by
an n-back working memory task. In the n-back task, participants were presented with a sequence of stimuli and asked to
indicate whether the current stimulus matched the one presented n steps before, thereby inducing a state of high
cognitive load \cite{jaeggiConcurrentValidityNBack2010}. Subsequently, participants completed a range of physical tasks
presented in randomized order: standing motionless in an upright posture; walking at a moderate pace (up to
$5\,\mathrm{km/h}$) with speed adjustments allowed for comfort; running at a brisker pace (up to $10\,\mathrm{km/h}$)
with speed modified for individual fitness and body characteristics; cycling on a stationary bike at a target speed of
$20\,\mathrm{km/h}$, with adjustments permitted; and lying down in four different positions right side, left side, back,
and prone position each for four minutes. The protocol concluded with a final four-minute seated period. Throughout,
procedures were tailored to ensure participant safety and comfort while preserving experimental rigor.

\subsection{Signal Quality Evaluation}
In this study, we conduct signal quality evaluations without the use of human annotation or subjective feedback by
domain experts. We assess the signal quality from multiple perspectives, each addressing distinct properties of the
acquired signals. Signal quality measures, commonly employed by related literature, are extracted to compare devices
during  identical physical activity types. Additionally, physiological features based on rhythmic patterns, are
extracted and compared between devices to further assess the integrity of the signals. We are accompanying this rhythm
based evaluation by conducting a statistical evaluation of beat-detection performance.
Morphological features, which hold significant clinical and pharmaceutical relevance, are also examined in this study.
Specifically, we analyze features such as QT intervals, commonly assessed in pharmaceutical research, as well as
amplitude markers associated with ventricular de- and repolarization. These amplitude markers are frequently utilized 
in clinical practice, for instance, in the detection of cardiac ischemia. 
Finally, we consider morphological changes that may intrinsically result from alternate electrode placement,
acknowledging that such variations can influence the overall signal characteristics. This multifaceted evaluation
strategy ensures a robust and objective assessment of signal quality.
Initial preprocessing involves signal filtering, followed by segmentation procedures, including R-peak detection and
fiducial point delineation. Detailed descriptions of these methods are provided in the Supplementary Material to ensure
reproducibility while maintaining focus on the main research objectives.
\subsubsection{Signal Quality Indices} We utilize \gls{sqis} to assess each acquisition device individually.
These \gls{sqis} are then compared across devices and tasks. Previous studies \cite{cliffordSignalQualityIndices2011,
beharSingleChannelECG2012, zhaoSQIQualityEvaluation2018, blasingECGPerformanceSimultaneous2022} introduced these
indices for evaluating the quality of ECG signals.
Table~\ref{tab:sqis} presents the names, descriptions, and corresponding references for each \gls{sqi}.
These indices capture multiple dimensions of signal quality, including morphological characteristics, spectral content,
and measures of signal regularity.
\begin{table}[]
    \caption{\gls{sqis} for ECG signals used in this study with references and a short description.}
    \label{tab:sqis}
    \centering
    \begin{tabular}{lll}
        \toprule
        Ref. & Name     & Description                                                                                                                                                   \\
        \midrule
        \cite{cliffordSignalQualityIndices2011}                      & sSQI     & Skewness of the signal                                                                                                                                                  \\
        \rowcolor[gray]{0.9} \cite{cliffordSignalQualityIndices2011} & kSQI     & Kurtosis of the signal \\ % https://pmc.ncbi.nlm.nih.gov/articles/PMC6011094/ kSQI > 5: optimal; kSQI <= 5 unqualified 
        \cite{beharSingleChannelECG2012}                             & pSQI     & \begin{tabular}[c]{@{}l@{}}Relative power in the QRS complex \\ $\left.\int_{5Hz}^{15Hz}P(f)df\!\middle/\,\int_{5Hz}^{40Hz}P(f)df\right.$ \end{tabular} \\ % $\frac{\int_{5Hz}^{15Hz}P(f)df}{\int_{5Hz}^{40Hz}P(f)df}$\end{tabular}                                          \\
        \rowcolor[gray]{0.9} \cite{haynQRSDetectionBased2012}                             & cSQI     & \begin{tabular}[c]{@{}l@{}}Regularity of RR intervals\end{tabular}                                                                                                              \\
        \cite{zhaoSQIQualityEvaluation2018}     & qSQI     & \begin{tabular}[c]{@{}l@{}}Matching degree of R peak detection\\ (Detectors: \cite{moodyXQRSAlgorithm2022} and  \cite{rodriguesLowComplexityRpeakDetection2021})\end{tabular}  \\
        \rowcolor[gray]{0.9} \cite{blasingECGPerformanceSimultaneous2022}                 & morphSQI & \begin{tabular}[c]{@{}l@{}}Morphological \gls{sqi} based on weighted cardiac \\cycle median deviation\end{tabular}                                               \\
        Ours                                    & vmSQI    & \begin{tabular}[c]{@{}l@{}}Mean absolute distance of complexes to the median \\ complex.\end{tabular}                              \\
        \bottomrule
    \end{tabular}
\end{table}
We calculate the \gls{sqis} for all leads and compare them across different phases. Previous studies have attempted to
establish thresholds based on heuristic assumptions or trained classifiers using expert annotations
on multiple \gls{sqis}.
However, because these heuristics and classifiers are typically tailored to specific tasks and rely on subjective
\gls{ecg} quality assessments, we calculate \gls{sqis} to enable an objective comparison between different 
acquisition devices.
% These SQIs have been introduced in previous studies as standardized metrics for assessing signal quality.
%
\subsubsection{Rhythmic Based Evaluation} 
\begin{figure}[ht!]
    \centering
    \includegraphics[]{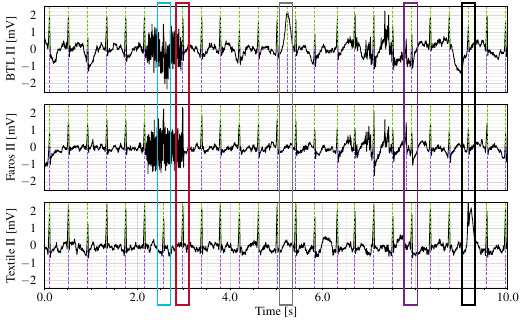}
    \caption{
Noisy data over time during an active session, showing signal artifacts and detection errors. Missed detections are
marked for BTL and Faros II (blue), Faros II (red), and Textile II (black); invalid detections in BTL II (grey); and
both invalid and missed detections in Faros II (purple). Movement and muscle noise are visible throughout. Dotted
lines indicate detection results from the Hamilton (green) \cite{hamiltonOpenSourceECG2002} and Rodrigues (purple) 
\cite{rodriguesLowComplexityRpeakDetection2021} detectors.}
    \label{fig:raw_ecg_detection_results}
    \vspace{-0.1cm}
\end{figure}

The second approach centers on the extraction and evaluation of widely
recognized features associated with rhythmic physiological changes, specifically \gls{hr} and \gls{hrv}. In line with
our initial methodology, we systematically compare these features across multiple acquisition devices, focusing on core
parameters such as \gls{hr}, the \gls{rmssd}, the \gls{sdnn}, and the \gls{hf} and \gls{lf} spectral components.
These metrics are extensively utilized in clinical contexts to assess autonomic function and cardiovascular health
\cite{shafferOverviewHeartRate2017}.
Given that the accurate estimation of these parameters depends on the reliable detection of heartbeats, typically
achieved by identifying R-peaks in the \gls{ecg} signal, we conduct a focused evaluation of R-peak detection
performance. Specifically, we evaluate detection quality across three distinct devices by selecting \gls{ecg} channel
positions analogous to lead II, which is commonly employed in peak detection tasks
\cite{cliffordAdvancedMethodsTools2006}. The classification of detection events is illustrated in
Figure~\ref{fig:raw_ecg_detection_results}, which provides a visual summary of typical signal artifacts and detection
errors encountered during an active recording session. The figure highlights various categories of errors, including
missed  detections and invalid detections across different device channels. Notably, the depicted interval exhibits
considerable movement and muscle noise artifacts.
To comprehensively assess detection performance, we compare a reference device with the textile-based system across
both male and female participants and throughout all experimental phases, accompanied by evaluating the reference
devices with each other. Both the BTL and Faros systems utilize Ag/AgCl electrodes positioned in proximity on the body,
thereby providing a robust reference measurement. To further elucidate the influence of signal projection, we report
hit, miss, and false alarm rates of the lead~I and V1 to the reference system Lead~II. This analysis underscores the
impact of electrode positioning and channel selection on detection fidelity.
To quantitatively compare the raw detection rates observed across different devices, we estimate the parameters of a
Dirichlet distribution using \gls{mle} \cite{minkaEstimatingDirichletDistribution2012}.  Probabilistic modeling of
detection events enables the performance of different devices to be represented by the parameter vector
$\boldsymbol{\alpha}$, facilitating a robust comparison between detection events of different devices.
\subsubsection{Classification}{} We trained classifiers on surrogate tasks to assess the predictive performance for
common \gls{ecg} applications using different acquisition devices. To accomplish this, we introduce two commonly
evaluated tasks in \gls{ecg} signal processing and train classifiers on the \gls{ecg} signals.
Both tasks employ rhythmic features, consistent with established applications in physiological signal processing of
\gls{ecg} signals. The \gls{hrv} features include time-domain measures, such as the \gls{sdnn} and the \gls{sdsd}.
Frequency-domain features are also utilized, specifically the low-frequency (0.04–0.15 Hz) and high-frequency
(0.15–0.4 Hz) components, as well as the ratio of high-to-low frequency spectral components (\gls{hf}/\gls{lf}). In
addition, features derived from the graphical Poincaré plot the SD1 and SD2 components and their ratio (SD1/SD2) are
incorporated. All features are extracted from lead~II.
Our activity recognition task is formulated as a three-class problem using a balanced dataset comprising the classes
\emph{Active}, \emph{Sitting}, and \emph{Lying}. The \emph{Active} class includes activities performed during
cycloergometer use, treadmill running, and treadmill walking. The \emph{Sitting} class encompasses periods when the
subject is sitting relaxed, watching a video, or participating in the n-back experiment. The \emph{Lying} class
captures instances where the subject is lying on their back, left side, or in prone position.
The second machine learning task utilizes the same feature set as input, but is designed to distinguish between two
classes reflecting different psycho-physiological states. The \emph{Low} load class represents a condition where the
subject is sitting upright at rest without any imposed task. In contrast, the \emph{High} load class corresponds to
the state in which the subject is engaged in solving the second level of the n-back test, a condition known 
to induce a high level of task load \cite{jaeggiConcurrentValidityNBack2010, oppeltADABaseMultimodalDataset2022}.
We trained our models using lead II data from both the primary reference system and the textile-based system. Feature
extraction was performed for each phase, followed by normalization via z-score transformation to achieve zero mean and 
unit standard deviation. For classification, we employed \emph{XGBoost}, a model widely used in feature-based detection
of psycho-physiological states \cite{oppeltADABaseMultimodalDataset2022}. The dataset was partitioned on a subject-wise
basis using a 5x5 nested cross-validation scheme. We report the mean and standard error of the \gls{auc} and $F_1$
score for both tasks and devices.
\subsubsection{Morphological Comparison}{} The morphological comparison of \gls{ecg} recordings obtained simultaneously
from two devices, each employing distinct electrode placements, presents inherent challenges due to the differing
spatial projections resulting from the cardiac electrical activity. Specifically, the orientation and placement of
electrodes significantly affect the measured signals, as each device records the cardiac electrical field from a
distinct spatial perspective. Integrating electrodes into a textile demands further adjustment of lead positions to
fulfill both ergonomic requirements and the minimization of noise artifacts due to relative electrode-skin movement, as
depicted in Figure~\ref{fig:positions}.
In addition, the use of the aforementioned silicon-based electrode material, in contrast to the Ag/AgCl wet
electrodes used in the reference system, exhibits distinct characteristics with respect to modeling the skin–electrode
interface \cite{yangInsightContactImpedance2022, goyalDependenceSkinElectrodeContact2022}.
These modifications are essential to ensure user comfort and robust signal acquisition but, in turn, complicate direct
morphological comparisons between signals derived from disparate lead configurations. Non-standard Einthoven placement 
of electrodes without detailed indication might cause diagnostic issues \cite{toosiFalseSTElevation2008}.
\begin{figure}[ht!]
\centering
    \includegraphics{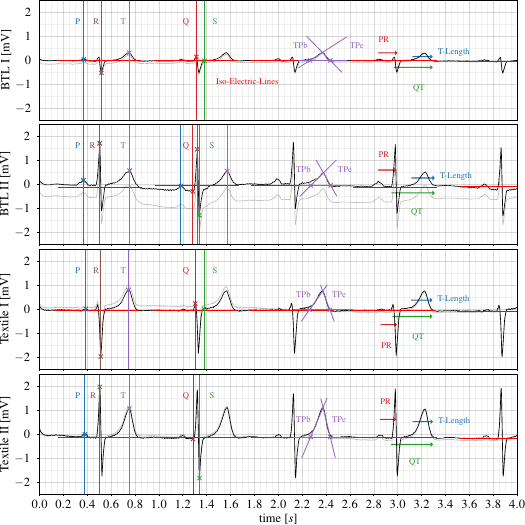}
\caption{
Lead I/II for BTL and textile devices, showing fiducial points (PQRST), isoelectric line, comparison intervals,
unfiltered (gray), and $0.67\,\mathrm{Hz}$ highpass-filtered (black) signals.} 
    \label{fig:morphology_sample}
    \vspace{-0.1cm}
\end{figure}
As an initial step in our morphological assessment, we extract commonly used fiducial points in \gls{ecg} analysis and
compute both interval- and amplitude-related measures. Specifically, we determine the amplitude from the isoelectric
line to the R wave, the amplitude of the ST segment, the amplitude from the isoelectric line to the T wave, and the
amplitude from the isoelectric line to the P wave. Additionally, we calculate the QT interval, representing the time
duration from the onset of the Q wave to the end of the T wave, as well as the T-length, defined as the interval between
T-onset and T-offset (extracted using tangents \cite{salviComparison5Methods2011}), as shown in
Figure~\ref{fig:morphology_sample}.
Direct projection or transformation of ECG signals between such configurations is nontrivial, as the inverse problem
is ill-posed and affected by factors such as respiratory motion, the variable composition of intervening tissues
(e.g., skin and fat layers), and, critically, the relative orientation of each electrode set with respect to the
cardiac vector \cite{liSolvingInverseProblem2025}. Recognizing these complexities, we approximate the correspondence
between the two electrode systems as a transformation comprising rotation and linear scaling within the lead I/II plane.
To formalize this, we frame the problem as a one-sided Procrustes analysis, where the objective is to
optimally align the median complexes extracted from the textile-based recordings, denoted as
$\mathbf{A} \in \mathbb{R}^{d \times n}$, to those from the reference BTL system, denoted as
$\mathbf{B} \in \mathbb{R}^{d \times n}$, with $d = 2$ corresponding to leads I and II and $n$ to the median complex
sequence length.
The transformation is modeled as a right-sided operation, with parameters determined by minimizing the squared
Frobenius norm $\|\cdot\|_F^2$, as shown in Equation~\ref{equ:procrustes_rotation}.
\begin{equation}
    \underbrace{\min}_{\left\{\mathbf{R} \left| {\mathbf{R}^{-1} = {\mathbf{R}}^\dagger
                                \atop \left| \mathbf{R} \right| = 1} \right. \right\}}
    \|\mathbf{A}\mathbf{R} - \mathbf{B}\|_{F}^2
    \label{equ:procrustes_rotation}
\end{equation}
%
% The solution to this optimization is obtained via the \gls{svd} of the cross-covariance matrix
% $\mathbf{A}^\dagger \mathbf{B}$.
To enhance alignment, we first normalize the textile-based median complex $\mathbf{A}$ to match the maximal amplitude
observed in the reference system $\mathbf{B}$. Specifically, for each sample in both $\mathbf{A}$ and $\mathbf{B}$, we
compute the instantaneous magnitude across leads I and II as $|\mathbf{v}| = \sqrt{I^2 + II^2}$. We then determine the
normalization constant $s$ as the ratio of the maximum absolute magnitude in the reference system to that of the textile
system, that is, $s = \max(|\mathbf{v}_B)|/|\max(\mathbf{v}_A|)$. The textile signals are subsequently scaled by this
constant prior to transformation, ensuring that both datasets are on a comparable amplitude scale and thereby improving
the robustness of the subsequent Procrustes alignment.
To quantitatively assess the quality of alignment between the transformed textile-based and reference \gls{ecg} signals,
we employ three similarity measures: the squared Frobenius norm error, the \gls{rmsd}, and the cosine similarity.
Upon determination of the optimal rotation parameters within the I/II plane, we further correlate these
findings with individual anthropometric measurements to elucidate which physiological attributes most significantly
influence the effects of altered electrode placement. This integrative approach not only enables rigorous morphological
comparison between textile and reference ECG recordings but also advances our understanding of the anthropometric
determinants impacting signal morphology under novel electrode configurations.
\section{Results}
\subsection{Comfort and Fit}
The subjects were provided with appropriately sized garments, with size distribution as follows: among male
participants, none wore size XS or XXL, while one wore size S, four wore size M, six wore size L, and four wore size XL. 
Among female participants, one wore size XS, ten wore size S, four wore size M, and none wore sizes L, XL, or XXL.
The wearable garments were designed with a unisex sizing scheme.
Participants were subsequently asked to assess their subjective comfort using a five-point Likert scale, ranging
from 1 (very uncomfortable, barely tolerable, and highly disturbing) to 5 (very comfortable, such that the wearer
forgot they were wearing the textile).
The overall comfort ratings for male participants were distributed as follows: two rated the textiles as 2, six as 3,
five as 4, and two as 5. For female participants, eight assigned a rating of 3, six a rating of 4, and one a rating of
5.
\subsection{Signal Quality}
\subsubsection{\gls{sqis}}
\begin{figure}[ht!]
\centering
    \includegraphics{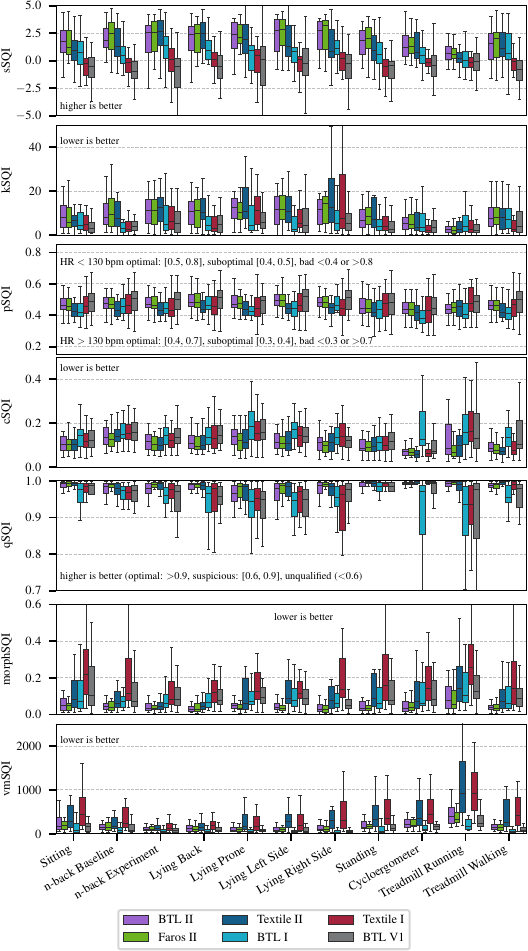}
    \caption{\gls{sqis} for all recorded leads and devices during different phases of the study with indication of
better signal quality according to original publication.}
    \label{fig:ecg_sqis}
    \vspace{-0.2cm}
\end{figure}
For \emph{cSQI}, the lowest values are observed in BTL, Faros, and Textile Lead II. Other leads not only exhibit higher
values during lying and sitting phases but also increase during more rigorous activities, such as treadmill running. 
Given that \emph{cSQI} is primarily based on RR interval detection, these findings suggest that the projection of
Lead~II is optimal for R-peak detection, with device-specific differences playing a minor role in this particular
\gls{sqi}.
The \emph{ksQI} highlights the pronounced effects of intense activities, with the lowest values recorded during
cycloergometer and treadmill running phases.
\emph{morphSQI} reveals the most pronounced changes across all \gls{sqis} for the textile electrode. This
observation underscores that the morphology of individual heart complexes can differ from the surrounding signal,
potentially due to baseline wander or other low-frequency variations.
Regarding \emph{pSQI}, although the median values remain similar across all phases, leads I and V1 exhibit greater
variance compared to lead II. This further emphasizes the impact of lead placement, which alters the ratio between
low and high frequency components in the QRS complexes.
The \emph{qSQI} demonstrates lower values during lying phases, while sitting and standing phases yield the highest
values. This index reflects the degree of concordance between two R-peak detectors, while lead~II of all systems
generally has higher values compared to lead~I and lead~V1. 
\emph{sSQI} shows that leads I and V1 consistently yield lower values than lead~II across all devices.
For \emph{vmSQI}, variance and changes are generally minimal across all leads, except during running and walking
phases, where increased variance is observed.
In summary, while several parameters suggest that lead II performs comparably to the reference systems even at
varying activity intensities, the statistical evaluation alone does not provide a comprehensive assessment. Lead
placement exerts a more significant influence on the computed \gls{sqis} than artifacts such as baseline wander, muscle
noise, or movement artifacts. Therefore, a thorough evaluation should prioritize practical applicability and intended
use rather than relying solely on statistical features.
\subsubsection{Physiological Parameters}
\begin{figure}[ht!]
    \centering
    \includegraphics[width=\linewidth]{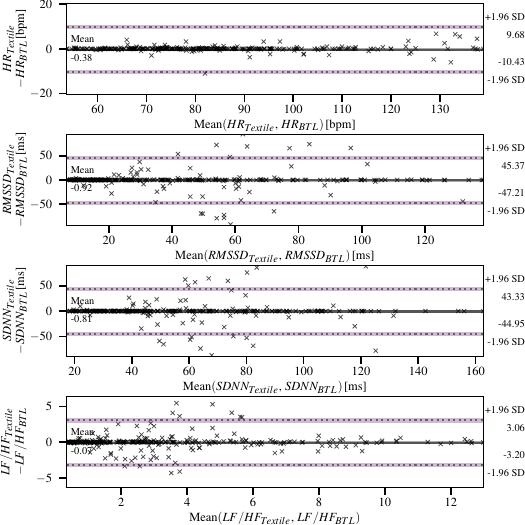}
    \caption{Bland-Altman plots comparing BTL reference and textile ECG systems for \gls{hr}, \gls{rmssd},
\gls{sdnn}, and \gls{hf}/\gls{lf}.}
    \label{fig:bland-altman-physiology}
\end{figure}

\begin{figure}[ht!]
    \centering
    \begin{subfigure}[t]{0.49\linewidth}
        \includegraphics[width=\textwidth]{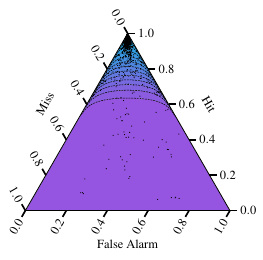}
        \caption{\centering \small BTL II vs. Textile II, $\boldsymbol{\alpha}=\left(11.96, 0.53, 0.52\right)$ }
        \label{sfig:btl-ecg-ii-cardiotextil-ecg-ii-dirichlet}
    \end{subfigure}
    \begin{subfigure}[t]{0.49\linewidth}
        \includegraphics[width=\textwidth]{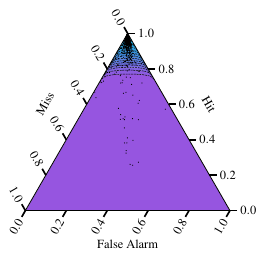}
        \caption{\centering \small BTL II vs. Faros II, $\boldsymbol{\alpha}=\left(29.01, 0.72, 0.70\right)$ }
        \label{sfig:btl-ecg-ii-faros-ecg-ii-dirichlet}
    \end{subfigure}
    \caption{
Log density of the Dirichlet distribution and parameter vector $\mathbf{\alpha}$ for miss, hit, and false alarm rates,
using BTL lead~II as reference and other leads for comparison.}
\label{fig:dirichlet_distribution}
\end{figure}

{} Mean \gls{hr}, as well as \gls{hrv} indices \gls{rmssd} and \gls{sdnn},
together with the \gls{lf}/\gls{hf} spectral ratio, are widely employed in fitness and psychophysiological assessments.
To evaluate the agreement between devices, we present these parameters using Bland-Altman plots, as illustrated in
Figure~\ref{fig:bland-altman-physiology}.
The results for \gls{hr} and time-domain \gls{hrv} measures show close agreement between devices for the majority of
measurements. The observed outliers are typically attributable to errors in R-peak detection. Specifically, the mean
difference is -0.38 for \gls{hr}, -0.93 for \gls{rmssd}, -0.86 for \gls{sdnn}, and -0.06 for the \gls{lf}/\gls{hf}
ratio, indicating a high level of concordance between the measurement devices.
No systematic errors are observed, such as higher values resulting in greater deviations between the two devices.
Visualizations in our supplementary material, reveal a high correlation of up to $R=0.93$, with minor reductions in
correlation primarily attributable to R-peak detection errors.
We present the hit rate, miss rate, and false alarm rate for all leads, compared to the reference device, across all
phases in Table~\ref{tab:sex_peak_performance}, separated by sex. The results indicate that the textile performs 
equally well for both female and male subjects, achieving a hit rate of $0.95\pm0.10$ for females and $0.95\pm0.12$
for males. This performance is comparable to that of the secondary reference device, Faros, which demonstrates hit
rates of $0.96\pm 0.07$ for females and $0.97\pm 0.04$ for males. Additionally, we report the R-peak detection
performance for lead~I, showing lower performance for females compared to males, indicating a strong impact of RA/LA
electrode placement on R-peak detectability.
\begin{table}[!ht]
    \renewcommand{\arraystretch}{1.3}
    \caption{R-peak detection performance by lead and sex, showing mean hit, miss, and false alarm rates.}
    \label{tab:sex_peak_performance}
      \centering
      \begin{NiceTabular}{l|l|ccc}
        \toprule
        ~ & ~ & Hit & Miss & False Alarm \\
        \midrule
        \multirow{5}{*}{Female} & BTL I & 0.89 $\pm$ 0.17 & 0.04 $\pm$ 0.08 & 0.07 $\pm$ 0.09 \\ 
        ~ &  BTL V1 & 0.80 $\pm$ 0.30 & 0.09 $\pm$ 0.15 & 0.11 $\pm$ 0.15 \\ 
        ~ &  Textile I & 0.74 $\pm$ 0.34 & 0.12 $\pm$ 0.17 & 0.14 $\pm$ 0.18 \\ 
        ~ &  Textile II & 0.95 $\pm$ 0.10 & 0.02 $\pm$ 0.05 & 0.03 $\pm$ 0.07 \\ 
        ~ &  Faros II & 0.96 $\pm$ 0.07 & 0.02 $\pm$ 0.03 & 0.02 $\pm$ 0.05 \\ 
        \midrule
        \multirow{5}{*}{Male} & BTL I & 0.94 $\pm$ 0.11 & 0.02 $\pm$ 0.04 & 0.04 $\pm$ 0.07 \\ 
        ~ &  BTL V1 & 0.76 $\pm$ 0.32 & 0.11 $\pm$ 0.16 & 0.13 $\pm$ 0.16 \\ 
        ~ &  Textile I & 0.90 $\pm$ 0.18 & 0.04 $\pm$ 0.08 & 0.06 $\pm$ 0.11 \\ 
        ~ &  Textile II & 0.95 $\pm$ 0.12 & 0.02 $\pm$ 0.04 & 0.03 $\pm$ 0.08 \\ 
        ~ &  Faros II & 0.97 $\pm$ 0.04 & 0.01 $\pm$ 0.01 & 0.02 $\pm$ 0.03 \\ 
        \bottomrule
    \end{NiceTabular}
\end{table}
As our recordings were neither annotated nor manually cleaned to remove noisy segments, the reference device is also
affected by reduced detection performance. The Faros reference device, based on the same technology, was positioned
in close-proximity to the BTL leads. Table~\ref{tab:three_device_confusion} presents the intersecting
R-peak detections among the three devices. Notably, $0.941$ of all R-peak events are detected by all devices, indicating
a substantial agreement. Conversely, $0.022$ of events are detected by both reference systems but not by the textile
electrode, providing a reliable estimate of the textile’s miss rate. Additionally, $0.007$ of events are not detected
by either the Faros or BTL device, which reflects potential false positives by the textile electrode.
When interpreting these results, it is essential to recognize that the reference systems were likely exposed to noise
sources similar to those affecting the textile device. Both reference systems utilized Ag/AgCl electrodes, which were
mounted in close proximity, closer than the textile electrodes, and connected by cables routed together. This
configuration increased their susceptibility to similar types of motion artifacts. In contrast, the textile device's
fundamentally different design may confer advantages that are not fully reflected in this configuration, potentially
leading to an underestimation of its performance. Therefore, this three-device comparison represents a conservative
assessment of the textile electrodes.
\begin{table}[!ht]
    \renewcommand{\arraystretch}{1.3}
    \caption{R-Peak detection performance of the two reference devices and the textile in lead~II.}
    \label{tab:three_device_confusion}
        \centering
        \begin{NiceTabular}{ll|C|C|ll}
        \toprule
    ~ & ~ & \Block{1-2}{BTL} & ~ & ~ \\ 
    ~ & ~ & detected & not det. & ~ & ~  \\ 
    \midrule
    \parbox[t]{2mm}{\multirow{4}{*}{\rotatebox[origin=c]{90}{Textile}}} & detected & $0.941 \pm 0.062$ & $0.004 \pm 0.01$ & \multirow{2}{*}{detected} & \parbox[t]{2mm}{\multirow{4}{*}{\rotatebox[origin=c]{90}{Faros}}} \\ 
    \cline{2-4} ~ & not det. & $0.002 \pm 0.003$ & $0.022 \pm 0.037$  & ~ & ~ \\
    \cline{2-5} ~ & detected & $0.016 \pm 0.026$ & $0.007 \pm 0.004$ & \multirow{2}{*}{not det.} & ~ \\ 
    \cline{2-4} ~ & not det. & $0.009 \pm 0.011$ & - & ~ & ~ \\ 
    \bottomrule
    \end{NiceTabular}
\end{table}
We estimate Dirichlet distributions for the observed events: hit, miss, and false alarm rates between pairs of devices
with the results of the estimated distributions presented in Figure~\ref{fig:dirichlet_distribution}. Both estimates
have a parameter configuration indicating that the majority of events are consistently identified by both devices.
To compare the relevant distributions, we compute the log-likelihood ratio between
$\mathcal{D}_{\mathrm{Textile~II}}^{\mathrm{BTL~II}}$ and $\mathcal{D}_{\mathrm{Faros~II}}^{\mathrm{BTL~II}}$,
obtaining a value of 0.03 during the resting phase. This result suggests that both distributions provide similar
likelihoods for the observed data. Additional results for other distributions and phases are provided in the
supplementary material.
\subsubsection{Classification}{}
The \gls{auc} and $F_1$ scores are presented in Table~\ref{tab:machine_learning_results}. For both tasks, the textile
electrode achieve predictive performance that is similar to, or even surpasses, that of machine learning models trained
on features extracted from lead~II of the reference system.
\begin{table}[!ht]
  \renewcommand{\arraystretch}{1.3}
  \caption{Machine Learning Results.}
  \label{tab:machine_learning_results}
      \centering
      \begin{NiceTabular}{l|cccccc}
      \toprule
~ & \multicolumn{2}{c}{Activity} &  \multicolumn{2}{c}{Psychophysiological}  \\
~ & AUC &  $F_1$ & AUC & $F_1$ \\
\midrule
BTL II & $0.88\pm0.03$ & $0.72\pm0.05$ & $0.84\pm0.05$ & $0.68\pm0.07$ \\
Textile II & $0.88\pm0.02$ & $0.70\pm0.04$ & $0.91\pm0.04$ & $0.72\pm0.05$ \\

\bottomrule
    \end{NiceTabular}
\end{table}
\subsubsection{Morphological Assessment}
\begin{figure}[ht!]
\centering
    \includegraphics{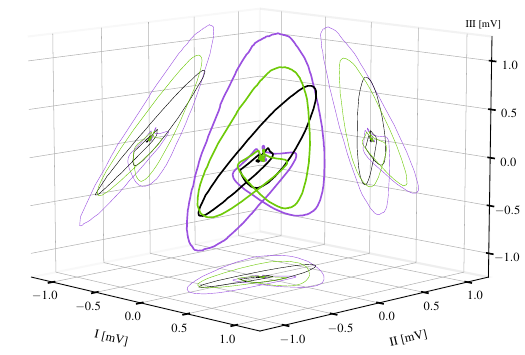}
\caption{3D heart vector: BTL leads I/II/III (black), textile original (purple), and rotated/scaled (green) with
2D-plane projections.}
    \label{fig:3d_heart}
    \vspace{-0.1cm}
\end{figure}

{}
The extracted amplitude and interval values follow approximately a normal distribution. Outliers were removed prior
to analysis. The Pearson correlation coefficient was computed between measurements from both devices for male and
female subjects separately. Despite the significant spatial separation of electrode projections, we observe strong
correlations between R-amplitudes, ST-amplitudes, and T-amplitudes, as well as between interval measures such as the
QT interval, across different leads and study phases as presented in Table~\ref{tab:morphological_features_correlation}.
Another finding is the pronounced influence of electrode positioning on amplitude values, particularly in the
supine position. The P-wave amplitude exhibits a lower correlation, as it was more challenging to consistently detect
the P-wave \cite{marsanovaBrnoUniversityTechnology2021}. Additionally, substantial sex differences are evident,
especially in the amplitude values of R-peaks for both lead~I and lead~II across all study phases indicate this impact.
\begin{table*}[!ht]
  \renewcommand{\arraystretch}{1.3}
  \caption{Pearson correlation $r$ of morphological features extracted from reference and system under test with 95\%CI
indicated as $\pm r$ {\tiny$\left[{\mathrm{CI}_{\mathrm{upper}} \atop \mathrm{CI}_{\mathrm{lower}}}\right]$} and
$\mathrm{abs}(r) > 0.3$ being in bold letters.}
  \label{tab:morphological_features_correlation}
    \centering
    \begin{NiceTabular}{l|l|l|rrrrrr}
      \toprule
 ~ & ~ & ~  & R-Ampl. & QT-Int. & ST-Ampl. & T-Ampl. & T-Length & P-Ampl. \\
      \midrule
 \parbox[t]{2mm}{\multirow{6}{*}{\rotatebox[origin=c]{90}{Female}}} &  \multirow{2}{*}{Sitting} &  BTL/Textile I  & $\mathbf{+0.65}$ \tiny{$\left[{ +0.66 \atop +0.63 }\right]$} & $\mathbf{+0.37}$ \tiny{$\left[{ +0.40 \atop +0.34 }\right]$} & $\mathbf{+0.53}$ \tiny{$\left[{ +0.55 \atop +0.51 }\right]$} & $\mathbf{+0.37}$ \tiny{$\left[{ +0.40 \atop +0.34 }\right]$} & $+0.18$ \tiny{$\left[{ +0.22 \atop +0.15 }\right]$} & $+0.18$ \tiny{$\left[{ +0.21 \atop +0.15 }\right]$} \\ 
 ~ & ~ & BTL/Textile II  & $\mathbf{+0.62}$ \tiny{$\left[{ +0.63 \atop +0.60 }\right]$} & $\mathbf{+0.46}$ \tiny{$\left[{ +0.49 \atop +0.43 }\right]$} & $\mathbf{+0.51}$ \tiny{$\left[{ +0.54 \atop +0.49 }\right]$} & $\mathbf{+0.57}$ \tiny{$\left[{ +0.59 \atop +0.55 }\right]$} & $+0.28$ \tiny{$\left[{ +0.31 \atop +0.24 }\right]$} & $\mathbf{+0.47}$ \tiny{$\left[{ +0.50 \atop +0.45 }\right]$} \\ 
 ~ & \multirow{2}{*}{Lying} &  BTL/Textile I  & $\mathbf{+0.42}$ \tiny{$\left[{ +0.45 \atop +0.40 }\right]$} & $\mathbf{+0.59}$ \tiny{$\left[{ +0.61 \atop +0.56 }\right]$} & $\mathbf{+0.63}$ \tiny{$\left[{ +0.65 \atop +0.61 }\right]$} & $+0.28$ \tiny{$\left[{ +0.31 \atop +0.26 }\right]$} & $+0.24$ \tiny{$\left[{ +0.27 \atop +0.21 }\right]$} & $+0.16$ \tiny{$\left[{ +0.19 \atop +0.13 }\right]$} \\ 
 ~ & ~ & BTL/Textile II  & $\mathbf{+0.32}$ \tiny{$\left[{ +0.35 \atop +0.29 }\right]$} & $\mathbf{+0.76}$ \tiny{$\left[{ +0.77 \atop +0.75 }\right]$} & $\mathbf{+0.42}$ \tiny{$\left[{ +0.44 \atop +0.40 }\right]$} & $\mathbf{+0.45}$ \tiny{$\left[{ +0.47 \atop +0.43 }\right]$} & $\mathbf{+0.55}$ \tiny{$\left[{ +0.58 \atop +0.53 }\right]$} & $\mathbf{+0.37}$ \tiny{$\left[{ +0.39 \atop +0.34 }\right]$} \\ 
 ~ & \multirow{2}{*}{Moving} &  BTL/Textile I  & $\mathbf{+0.68}$ \tiny{$\left[{ +0.69 \atop +0.66 }\right]$} & $\mathbf{+0.43}$ \tiny{$\left[{ +0.46 \atop +0.41 }\right]$} & $\mathbf{+0.42}$ \tiny{$\left[{ +0.44 \atop +0.39 }\right]$} & $+0.17$ \tiny{$\left[{ +0.20 \atop +0.14 }\right]$} & $+0.28$ \tiny{$\left[{ +0.31 \atop +0.25 }\right]$} & $+0.10$ \tiny{$\left[{ +0.13 \atop +0.07 }\right]$} \\ 
 ~ & ~ & BTL/Textile II  & $\mathbf{+0.48}$ \tiny{$\left[{ +0.50 \atop +0.45 }\right]$} & $\mathbf{+0.36}$ \tiny{$\left[{ +0.39 \atop +0.33 }\right]$} & $\mathbf{+0.50}$ \tiny{$\left[{ +0.52 \atop +0.47 }\right]$} & $+0.28$ \tiny{$\left[{ +0.31 \atop +0.25 }\right]$} & $+0.23$ \tiny{$\left[{ +0.26 \atop +0.20 }\right]$} & $+0.19$ \tiny{$\left[{ +0.22 \atop +0.16 }\right]$} \\ 
 \midrule 
 \parbox[t]{2mm}{\multirow{6}{*}{\rotatebox[origin=c]{90}{Male}}} &  \multirow{2}{*}{Sitting} &  BTL/Textile I  & $\mathbf{+0.80}$ \tiny{$\left[{ +0.81 \atop +0.79 }\right]$} & $\mathbf{+0.45}$ \tiny{$\left[{ +0.47 \atop +0.43 }\right]$} & $\mathbf{+0.67}$ \tiny{$\left[{ +0.69 \atop +0.66 }\right]$} & $\mathbf{+0.40}$ \tiny{$\left[{ +0.42 \atop +0.38 }\right]$} & $+0.21$ \tiny{$\left[{ +0.23 \atop +0.18 }\right]$} & $\mathbf{+0.31}$ \tiny{$\left[{ +0.34 \atop +0.29 }\right]$} \\ 
 ~ & ~ & BTL/Textile II  & $+0.18$ \tiny{$\left[{ +0.21 \atop +0.16 }\right]$} & $\mathbf{+0.36}$ \tiny{$\left[{ +0.38 \atop +0.34 }\right]$} & $\mathbf{+0.59}$ \tiny{$\left[{ +0.61 \atop +0.57 }\right]$} & $\mathbf{+0.43}$ \tiny{$\left[{ +0.45 \atop +0.41 }\right]$} & $+0.08$ \tiny{$\left[{ +0.10 \atop +0.05 }\right]$} & $+0.13$ \tiny{$\left[{ +0.16 \atop +0.10 }\right]$} \\ 
 ~ & \multirow{2}{*}{Lying} &  BTL/Textile I  & $\mathbf{+0.75}$ \tiny{$\left[{ +0.76 \atop +0.74 }\right]$} & $\mathbf{+0.36}$ \tiny{$\left[{ +0.38 \atop +0.34 }\right]$} & $\mathbf{+0.41}$ \tiny{$\left[{ +0.43 \atop +0.39 }\right]$} & $\mathbf{+0.38}$ \tiny{$\left[{ +0.40 \atop +0.36 }\right]$} & $+0.09$ \tiny{$\left[{ +0.12 \atop +0.07 }\right]$} & $\mathbf{+0.38}$ \tiny{$\left[{ +0.40 \atop +0.36 }\right]$} \\ 
 ~ & ~ & BTL/Textile II  & $\mathbf{-0.39}$ \tiny{$\left[{ -0.37 \atop -0.41 }\right]$} & $\mathbf{+0.57}$ \tiny{$\left[{ +0.58 \atop +0.55 }\right]$} & $\mathbf{+0.35}$ \tiny{$\left[{ +0.37 \atop +0.32 }\right]$} & $+0.26$ \tiny{$\left[{ +0.28 \atop +0.24 }\right]$} & $+0.11$ \tiny{$\left[{ +0.13 \atop +0.08 }\right]$} & $+0.02$ \tiny{$\left[{ +0.04 \atop -0.01 }\right]$} \\ 
 ~ & \multirow{2}{*}{Moving} &  BTL/Textile I  & $\mathbf{+0.75}$ \tiny{$\left[{ +0.76 \atop +0.74 }\right]$} & $+0.24$ \tiny{$\left[{ +0.26 \atop +0.22 }\right]$} & $\mathbf{+0.44}$ \tiny{$\left[{ +0.46 \atop +0.42 }\right]$} & $+0.14$ \tiny{$\left[{ +0.16 \atop +0.11 }\right]$} & $+0.13$ \tiny{$\left[{ +0.16 \atop +0.10 }\right]$} & $+0.12$ \tiny{$\left[{ +0.15 \atop +0.10 }\right]$} \\ 
 ~ & ~ & BTL/Textile II  & $-0.09$ \tiny{$\left[{ -0.07 \atop -0.12 }\right]$} & $+0.24$ \tiny{$\left[{ +0.27 \atop +0.22 }\right]$} & $+0.24$ \tiny{$\left[{ +0.26 \atop +0.22 }\right]$} & $+0.08$ \tiny{$\left[{ +0.10 \atop +0.05 }\right]$} & $+0.11$ \tiny{$\left[{ +0.14 \atop +0.09 }\right]$} & $+0.05$ \tiny{$\left[{ +0.07 \atop +0.02 }\right]$} \\
      \bottomrule
    \end{NiceTabular}
\end{table*}
To further study the effect of this pronounced sex difference we rotate and scale the textile recordings as shown in
Figure~\ref{fig:3d_heart}.
The results of the alignment procedure, presented in Table~\ref{tab:quality_procrustes_alignment}, demonstrate
improved cosine similarity $S_c$ and reduced Frobenius distance after alignment, indicating enhanced correspondence between
the recordings.
\begin{table}[!ht]
  \renewcommand{\arraystretch}{1.3}
  \caption{Quality of Procrustes Alignment.}
  \label{tab:quality_procrustes_alignment}
      \centering
      \begin{NiceTabular}{l|ccc}
      \toprule
~ & RMSD &  $S_C$ & $\|A\|_F^2$ \\
\midrule
Original Textile \& BTL & $0.14\pm0.05$ & $0.76\pm0.08$ & $64.15\pm40.7$ \\
Rotated Textile \& BTL & $0.06\pm0.02$ & $0.78\pm0.07$ & $11.39\pm9.01$ \\
\bottomrule
    \end{NiceTabular}
\end{table}
\begin{table}[!ht]
  \renewcommand{\arraystretch}{1.2}
  \caption{Pearson correlation $r$ of rotation angles and anthropometric measurements with 95\%\,CI.}
  \label{tab:procrustes_alginment_correlations}
    \centering
    \begin{NiceTabular}{L|RR}
      \toprule
 ~ & Female Rotation~I/II & Male Rotation~I/II \\
      \midrule
z1r & $-0.26$ \tiny{$\left[{ +0.02 \atop -0.47 }\right]$} & $-0.09$ \tiny{$\left[{ +0.18 \atop -0.33 }\right]$} \\ 
z2r & $-0.16$ \tiny{$\left[{ +0.11 \atop -0.39 }\right]$} & $-0.09$ \tiny{$\left[{ +0.18 \atop -0.33 }\right]$} \\ 
z1l & $-0.24$ \tiny{$\left[{ +0.03 \atop -0.46 }\right]$} & $-0.19$ \tiny{$\left[{ +0.08 \atop -0.42 }\right]$} \\ 
z2l & $+0.14$ \tiny{$\left[{ +0.38 \atop -0.13 }\right]$} & $-0.08$ \tiny{$\left[{ +0.18 \atop -0.33 }\right]$} \\ 
z1m & $\mathbf{-0.56}$ \tiny{$\left[{ -0.32 \atop -0.70 }\right]$} & $+0.23$ \tiny{$\left[{ +0.44 \atop -0.05 }\right]$} \\ 
z2m & $\mathbf{-0.45}$ \tiny{$\left[{ -0.18 \atop -0.61 }\right]$} & $-0.14$ \tiny{$\left[{ +0.13 \atop -0.38 }\right]$} \\ 
z3r & $\mathbf{-0.32}$ \tiny{$\left[{ -0.04 \atop -0.51 }\right]$} & $-0.21$ \tiny{$\left[{ +0.07 \atop -0.43 }\right]$} \\ 
z3l & $-0.24$ \tiny{$\left[{ +0.04 \atop -0.45 }\right]$} & $\mathbf{-0.42}$ \tiny{$\left[{ -0.15 \atop -0.60 }\right]$} \\ 
a1r & $-0.29$ \tiny{$\left[{ -0.02 \atop -0.50 }\right]$} & $-0.12$ \tiny{$\left[{ +0.15 \atop -0.36 }\right]$} \\ 
a2r & $\mathbf{-0.41}$ \tiny{$\left[{ -0.14 \atop -0.59 }\right]$} & $-0.01$ \tiny{$\left[{ +0.25 \atop -0.26 }\right]$} \\ 
a1l & $-0.22$ \tiny{$\left[{ +0.06 \atop -0.44 }\right]$} & $-0.12$ \tiny{$\left[{ +0.15 \atop -0.35 }\right]$} \\ 
a2l & $\mathbf{-0.46}$ \tiny{$\left[{ -0.20 \atop -0.63 }\right]$} & $\mathbf{+0.44}$ \tiny{$\left[{ +0.61 \atop +0.18 }\right]$} \\ 
Height & $\mathbf{+0.39}$ \tiny{$\left[{ +0.57 \atop +0.12 }\right]$} & $-0.08$ \tiny{$\left[{ +0.18 \atop -0.33 }\right]$} \\ 
Weight & $-0.14$ \tiny{$\left[{ +0.13 \atop -0.38 }\right]$} & $-0.20$ \tiny{$\left[{ +0.07 \atop -0.43 }\right]$} \\
      \bottomrule
    \end{NiceTabular}
\end{table}
The analysis of textile electrode placement as shown in Figure~\ref{sfig:electrode-measurements} and rotation angle
reveals distinct sex-specific correlations. Among female participants, significant associations are observed between
the rotation angle and the distances from both the upper (z1m) and lower (z2m) textile electrodes to the reference
electrodes. Additionally, the distances from the upper reference electrode to the upper right (a2r) and upper left (a2l)
textile electrodes also demonstrate significant correlations. Body height emerges as a notable influencing factor in
females, whereas such correlations are considerably weaker in males.
Conversely, in male participants, significant correlations with rotation are found for the distances to the upper
reference electrodes, particularly z3l and a2l. Distances to the axillary line further contribute to the observed
patterns in males.
Collectively, these results highlight sex-specific differences in the factors affecting textile electrode placement and
its relationship with biophysical measurements. In particular, for females, who generally have smaller body
dimensions, the width and height of the upper body substantially impact \gls{ecg} signal formation in textile electrode
systems. This effect is especially important when electrodes are positioned closer to the body center, compared
to the standard Einthoven placement.
\section{Conclusion}
Our study presents a comprehensive assessment of \gls{ecg} signal quality obtained from a wearable garment with
textile-based dry electrodes, benchmarked against gold-standard Holter \gls{ecg} systems. Our evaluation encompassed
multiple dimensions, including \gls{sqis}, rhythm-based parameters, classification tasks relevant to
physiological and psychophysiological states, and detailed morphological comparisons.

The results demonstrate that the textile-based system achieves concordance with reference devices in \gls{hr}
and \gls{hrv} metrics across diverse activities and postures. Analysis of \gls{sqis} demonstrates that lead
placement significantly influences signal quality. The textile system achieves signal quality comparable to medical
grade reference devices, especially in the commonly used lead~II. Machine learning-based classification tasks further
confirms that features extracted from the textile system enable robust activity and task load discrimination,
matching the performance of features from reference system recordings.
Morphological analyzes underline strong correlations between key \gls{ecg} parameters recorded by textile and reference
systems, although spatial and anatomical variations in electrode placement can affect amplitude values, most notably
in the supine position and among female participants. Our Procrustes alignment approach, incorporating rotation, 
enhances the morphological correspondence between systems and enables a more nuanced interpretation of inter-device
differences. Notably, the investigation of electrode placement and anthropometric correlations revealed sex-specific
determinants: especially for females, the distances between upper and lower textile electrodes and height are principal
factors affecting rotation.

These findings collectively underscore the viability of textile electrode garments for long-term, ambulatory \gls{ecg}
monitoring, offering signal quality and diagnostic utility comparable to clinical gold standards. Importantly, the study
highlights the necessity to account for sex-specific anatomical differences in garment design and electrode placement
to ensure equitable signal quality and diagnostic accuracy.

This work advances the evaluation of novel wearable textile \gls{ecg} acquisition devices by introducing a rigorous,
multi-faceted assessment framework. It comprehensively examines signal quality, usability in real-world applications,
and incorporates sex-specific anatomical differences as well as key ergonomic and physiological determinants that
inform the development of next-generation wearable \gls{ecg} systems.

\subsection{Limitations}
Despite the robust design and comprehensive evaluation framework employed in this study, several limitations should be
acknowledged. The electrode placement on the textile garment, although intended to closely approximate standardized
locations, is inherently affected by anatomical variability and differences in garment fit among participants. These
factors may influence signal quality and comparability with the reference systems. Although the cohort was balanced by
sex and included a range of body types, the sample size may still limit the generalizability of findings, particularly
regarding specific subpopulations or individuals with pathological cardiac conditions.
Furthermore, all measurements were conducted under controlled laboratory conditions and predominantly involved healthy
subjects; thus, performance under real-world ambulatory settings or in patient populations remains to be validated.
Additionally, while objective \gls{sqis} and automated analysis were prioritized to minimize subjective bias, the lack
of expert manual annotation may have led to an underestimation of device-specific artifacts or subtle diagnostic
discrepancies. Finally, the scope of morphological comparisons is constrained by non-standardized lead projections,
complicating direct equivalence between textile and clinical reference recordings. Future studies should address these
limitations by incorporating larger and more diverse cohorts, testing in varied ambulatory environments, and including
expert-annotated data for comprehensive clinical validation.

\bibliographystyle{IEEEtran}
\bibliography{bibliography}

\begin{thebibliography}{10}
\providecommand{\url}[1]{#1}
\csname url@samestyle\endcsname
\providecommand{\newblock}{\relax}
\providecommand{\bibinfo}[2]{#2}
\providecommand{\BIBentrySTDinterwordspacing}{\spaceskip=0pt\relax}
\providecommand{\BIBentryALTinterwordstretchfactor}{4}
\providecommand{\BIBentryALTinterwordspacing}{\spaceskip=\fontdimen2\font plus
\BIBentryALTinterwordstretchfactor\fontdimen3\font minus
  \fontdimen4\font\relax}
\providecommand{\BIBforeignlanguage}[2]{{%
\expandafter\ifx\csname l@#1\endcsname\relax
\typeout{** WARNING: IEEEtran.bst: No hyphenation pattern has been}%
\typeout{** loaded for the language `#1'. Using the pattern for}%
\typeout{** the default language instead.}%
\else
\language=\csname l@#1\endcsname
\fi
#2}}
\providecommand{\BIBdecl}{\relax}
\BIBdecl

\bibitem{einthovenGalvanometrischeRegistrirungMenschlichen1903}
W.~Einthoven, ``{Die galvanometrische Registrirung des menschlichen
  Elektrokardiogramms, zugleich eine Beurtheilung der Anwendung des
  Capillar-Elektrometers in der Physiologie},'' \emph{Physiologie des Menschen
  und der Tiere}, vol.~99, no. 9-10, Nov. 1903.

\bibitem{kligfieldRecommendationsStandardizationInterpretation2007}
P.~Kligfield, L.~S. Gettes, J.~J. Bailey, R.~Childers, B.~J. Deal, E.~W.
  Hancock, G.~{van Herpen}, J.~A. Kors, P.~Macfarlane, D.~M. Mirvis, O.~Pahlm,
  P.~Rautaharju, and G.~S. Wagner, ``Recommendations for the
  {{Standardization}} and {{Interpretation}} of the {{Electrocardiogram}}:
  {{Part I}}: {{The Electrocardiogram}} and {{Its Technology}}: {{A Scientific
  Statement From}} the {{American Heart Association Electrocardiography}} and
  {{Arrhythmias Committee}}, {{Council}} on {{Clinical Cardiology}}; the
  {{American College}} of {{Cardiology Foundation}}; and the {{Heart Rhythm
  Society}} {{{\emph{Endorsed}}}}{\emph{ by the }}{{{\emph{International
  Society}}}}{\emph{ for }}{{{\emph{Computerized Electrocardiology}}}},''
  \emph{Circulation}, vol. 115, no.~10, Mar. 2007.

\bibitem{reisnerPhysiologicalBasisElectrocardiogram2006}
A.~T. Reisner, G.~D. Clifford, and R.~G. Mark, ``The {{Physiological Basis}} of
  the {{Electrocardiogram}},'' in \emph{Advanced {{Methods}} and {{Tools}} for
  {{ECG Data Analysis}}}, ser. Engineering in {{Medicine}} \&
  {{Biology}}.\hskip 1em plus 0.5em minus 0.4em\relax Boston, London: Artech
  House, 2006.

\bibitem{masonRecommendationsStandardizationInterpretation2007}
J.~W. Mason, E.~W. Hancock, and L.~S. Gettes, ``Recommendations for the
  {{Standardization}} and {{Interpretation}} of the {{Electrocardiogram}}:
  {{Part II}}: {{Electrocardiography Diagnostic Statement List}}: {{A
  Scientific Statement From}} the {{American Heart Association
  Electrocardiography}} and {{Arrhythmias Committee}}, {{Council}} on
  {{Clinical Cardiology}}; the {{American College}} of {{Cardiology
  Foundation}}; and the {{Heart Rhythm Society}}: {{{\emph{Endorsed}}}}{\emph{
  by the }}{{{\emph{International Society}}}}{\emph{ for
  }}{{{\emph{Computerized Electrocardiology}}}},'' \emph{Circulation}, vol.
  115, no.~10, Mar. 2007.

\bibitem{sternDiagnosticAccuracyAmbulatory1975}
S.~Stern, D.~Tzivoni, and Z.~Stern, ``Diagnostic accuracy of ambulatory {{ECG}}
  monitoring in ischemic heart disease.'' \emph{Circulation}, vol.~52, no.~6,
  Dec. 1975.

\bibitem{eichhornStandardsPatientMonitoring1986}
J.~H. Eichhorn, ``Standards for {{Patient Monitoring During Anesthesia}} at
  {{Harvard Medical School}},'' \emph{JAMA: The Journal of the American Medical
  Association}, vol. 256, no.~8, Aug. 1986.

\bibitem{malikEvaluationDrugInducedQT2001}
M.~Malik and A.~J. Camm, ``Evaluation of {{Drug-Induced QT Interval
  Prolongation}}: {{Implications}} for {{Drug Approval}} and {{Labelling}},''
  \emph{Drug Safety}, vol.~24, no.~5, 2001.

\bibitem{demmelNoQTcProlongation2018}
V.~Demmel, A.~{Sandberg-Schaal}, J.~B. Jacobsen, G.~Golor, J.~Pettersson, and
  A.~Flint, ``No {{QTc Prolongation}} with {{Semaglutide}}: {{A Thorough QT
  Study}} in {{Healthy Subjects}},'' \emph{Diabetes Therapy}, vol.~9, no.~4,
  Aug. 2018.

\bibitem{oppeltADABaseMultimodalDataset2022}
M.~P. Oppelt, A.~Foltyn, J.~Deuschel, N.~R. Lang, N.~Holzer, B.~M. Eskofier,
  and S.~H. Yang, ``{{ADABase}}: {{A Multimodal Dataset}} for {{Cognitive Load
  Estimation}},'' \emph{Sensors}, vol.~23, no.~1, p. 340, Dec. 2022.

\bibitem{sahaInvestigationRelationPhysiological2021}
B.~Saha, L.~Becker, J.-U. Garbas, M.~Oppelt, A.~Foltyn, S.~Hettenkofer,
  N.~Lang, M.~Struck, N.~Rohleder, and B.~Mahesh, ``Investigation of
  {{Relation}} between {{Physiological Responses}} and {{Personality}} during
  {{Stress Recovery}},'' in \emph{2021 {{IEEE International Conference}} on
  {{Pervasive Computing}} and {{Communications Workshops}} and Other
  {{Affiliated Events}} ({{PerCom Workshops}})}.\hskip 1em plus 0.5em minus
  0.4em\relax Kassel, Germany: IEEE, Mar. 2021.

\bibitem{hanselContactSensitivityElectrocardiogram2020}
K.~Hansel, M.~Tramontana, L.~Bianchi, E.~Cerulli, C.~Patruno, M.~Napolitano,
  and L.~Stingeni, ``Contact sensitivity to electrocardiogram electrodes due to
  acrylic acid: {{A}} rare cause of medical device allergy,'' \emph{Contact
  Dermatitis}, vol.~82, no.~2, Feb. 2020.

\bibitem{fotiContactAllergyElectrocardiogram2018}
C.~Foti, A.~Lopalco, L.~Stingeni, K.~Hansel, A.~Lopedota, N.~Denora, and
  P.~Romita, ``Contact allergy to electrocardiogram electrodes caused by
  acrylic acid without sensitivity to methacrylates and ethyl cyanoacrylate,''
  \emph{Contact Dermatitis}, vol.~79, no.~2, Jul. 2018.

\bibitem{crawfordPracticalAspectsECG2012}
J.~Crawford and L.~Doherty, \emph{Practical Aspects of {{ECG}}
  Recording}.\hskip 1em plus 0.5em minus 0.4em\relax Cumbria [England]: M \& K
  Update, 2012.

\bibitem{evensonReviewValidityReliability2020}
K.~R. Evenson and C.~L. Spade, ``Review of {{Validity}} and {{Reliability}} of
  {{Garmin Activity Trackers}},'' \emph{Journal for the Measurement of Physical
  Behaviour}, vol.~3, no.~2, Jun. 2020.

\bibitem{caoAccuracyAssessmentOura2022}
R.~Cao, I.~Azimi, F.~Sarhaddi, H.~{Niela-Vilen}, A.~Axelin, P.~Liljeberg, and
  A.~M. Rahmani, ``Accuracy {{Assessment}} of {{Oura Ring Nocturnal Heart
  Rate}} and {{Heart Rate Variability}} in {{Comparison With
  Electrocardiography}} in {{Time}} and {{Frequency Domains}}: {{Comprehensive
  Analysis}},'' \emph{Journal of Medical Internet Research}, vol.~24, no.~1,
  Jan. 2022.

\bibitem{el-amrawyAreCurrentlyAvailable2015}
F.~{El-Amrawy} and M.~I. Nounou, ``Are {{Currently Available Wearable Devices}}
  for {{Activity Tracking}} and {{Heart Rate Monitoring Accurate}},
  {{Precise}}, and {{Medically Beneficial}}?'' \emph{Healthcare Informatics
  Research}, vol.~21, no.~4, 2015.

\bibitem{peakeCriticalReviewConsumer2018}
J.~M. Peake, G.~Kerr, and J.~P. Sullivan, ``A {{Critical Review}} of {{Consumer
  Wearables}}, {{Mobile Applications}}, and {{Equipment}} for {{Providing
  Biofeedback}}, {{Monitoring Stress}}, and {{Sleep}} in {{Physically Active
  Populations}},'' \emph{Frontiers in Physiology}, vol.~9, Jun. 2018.

\bibitem{fineSourcesInaccuracyPhotoplethysmography2021}
J.~Fine, K.~L. Branan, A.~J. Rodriguez, T.~{Boonya-ananta}, {Ajmal}, J.~C.
  {Ramella-Roman}, M.~J. McShane, and G.~L. Cot{\'e}, ``Sources of
  {{Inaccuracy}} in {{Photoplethysmography}} for {{Continuous Cardiovascular
  Monitoring}},'' \emph{Biosensors}, vol.~11, no.~4, p. 126, Apr. 2021.

\bibitem{aneeshs.FDAElectrocardiographSoftware2020}
D.~S. Aneesh~S. and J.~Kozen~Shih, ``{{FDA Electrocardiograph Software}} for
  {{Over-The-Counter Use Regulatory Class}}: {{Class II}},'' 2020.

\bibitem{haverkampVorhofflimmerndiagnostikMittelsEKGfaehiger2022}
W.~Haverkamp, O.~G{\"o}ing, M.~Anker, and S.~D. Anker,
  ``{Vorhofflimmerndiagnostik mittels EKG-f{\"a}higer Smartwatches},''
  \emph{Der Nervenarzt}, vol.~93, no.~2, Feb. 2022.

\bibitem{veltmannWearablebasierteDetektionArrhythmien2021}
C.~Veltmann, J.~R. Ehrlich, U.~M. Gassner, B.~Meder, M.~M{\"o}ckel, P.~Radke,
  E.~Scholz, H.~Schneider, C.~Stellbrink, and D.~Duncker, ``{Wearable-basierte
  Detektion von Arrhythmien},'' \emph{Der Kardiologe}, vol.~15, no.~4, pp.
  341--353, Aug. 2021.

\bibitem{bhargavaAliveCor2018}
B.~Bhargava, ``{{AliveCor}},'' \emph{Journal of the Practice of Cardiovascular
  Sciences}, vol.~4, no.~1, 2018.

\bibitem{halcoxAssessmentRemoteHeart2017}
J.~P. Halcox, K.~Wareham, A.~Cardew, M.~Gilmore, J.~P. Barry, C.~Phillips, and
  M.~B. Gravenor, ``Assessment of {{Remote Heart Rhythm Sampling Using}} the
  {{AliveCor Heart Monitor}} to {{Screen}} for {{Atrial Fibrillation}}: {{The
  REHEARSE-AF Study}},'' \emph{Circulation}, vol. 136, no.~19, Nov. 2017.

\bibitem{chungQTCIntervalsCan2015}
E.~H. Chung and K.~D. Guise, ``{{QTC}} intervals can be assessed with the
  {{AliveCor}} heart monitor in patients on dofetilide for atrial
  fibrillation,'' \emph{Journal of Electrocardiology}, vol.~48, no.~1, Jan.
  2015.

\bibitem{nigusseWearableSmartTextiles2021}
A.~B. Nigusse, D.~A. Mengistie, B.~Malengier, G.~B. Tseghai, and L.~V.
  Langenhove, ``Wearable {{Smart Textiles}} for {{Long-Term Electrocardiography
  Monitoring- A Review}},'' \emph{Sensors}, vol.~21, no.~12, p. 4174, Jun.
  2021.

\bibitem{alizadeh-meghraziEvaluationDryTextile2021}
M.~{Alizadeh-Meghrazi}, B.~Ying, A.~Schlums, E.~Lam, L.~Eskandarian, F.~Abbas,
  G.~Sidhu, A.~Mahnam, B.~Moineau, and M.~R. Popovic, ``Evaluation of dry
  textile electrodes for long-term electrocardiographic monitoring,''
  \emph{BioMedical Engineering OnLine}, vol.~20, no.~1, Jul. 2021.

\bibitem{eskandarianRobustMultifunctionalConductive2020}
L.~Eskandarian, E.~Lam, C.~Rupnow, M.~A. Meghrazi, and H.~E. Naguib, ``Robust
  and {{Multifunctional Conductive Yarns}} for {{Biomedical Textile
  Computing}},'' \emph{ACS Applied Electronic Materials}, vol.~2, no.~6, pp.
  1554--1566, Jun. 2020.

\bibitem{acarWearableFlexibleTextile2019}
G.~Acar, O.~Ozturk, A.~J. Golparvar, T.~A. Elboshra, K.~B{\"o}hringer, and
  M.~K. Yapici, ``Wearable and {{Flexible Textile Electrodes}} for
  {{Biopotential Signal Monitoring}}: {{A}} review,'' \emph{Electronics},
  vol.~8, no.~5, p. 479, Apr. 2019.

\bibitem{catrysseIntegrationTextileSensors2004}
M.~Catrysse, R.~Puers, C.~Hertleer, L.~Van~Langenhove, H.~Van~Egmond, and
  D.~Matthys, ``Towards the integration of textile sensors in a wireless
  monitoring suit,'' \emph{Sensors and Actuators A: Physical}, vol. 114, no.
  2-3, pp. 302--311, Sep. 2004.

\bibitem{marquezComparisonDrytextileElectrodes2010}
J.~C. M{\'a}rquez, F.~Seoane, E.~V{\"a}lim{\"a}ki, and K.~Lindecrantz,
  ``Comparison of dry-textile electrodes for electrical bioimpedance
  spectroscopy measurements,'' \emph{Journal of Physics: Conference Series},
  vol. 224, p. 012140, Apr. 2010.

\bibitem{yooWearableECGAcquisition2009}
J.~Yoo, {Long Yan}, {Seulki Lee}, {Hyejung Kim}, and {Hoi-Jun Yoo}, ``A
  {{Wearable ECG Acquisition System With Compact Planar-Fashionable Circuit
  Board-Based Shirt}},'' \emph{IEEE Transactions on Information Technology in
  Biomedicine}, vol.~13, no.~6, Nov. 2009.

\bibitem{tseghaiIntegrationConductiveMaterials2020}
G.~B. Tseghai, B.~Malengier, K.~A. Fante, A.~B. Nigusse, and L.~Van~Langenhove,
  ``Integration of {{Conductive Materials}} with {{Textile Structures}}, an
  {{Overview}},'' \emph{Sensors}, vol.~20, no.~23, p. 6910, Dec. 2020.

\bibitem{paniSurveyTextileElectrode2018}
D.~Pani, A.~Achilli, and A.~Bonfiglio, ``Survey on {{Textile Electrode
  Technologies}} for {{Electrocardiographic}} ({{ECG}}) {{Monitoring}}, from
  {{Metal Wires}} to {{Polymers}},'' \emph{Advanced Materials Technologies},
  vol.~3, no.~10, pp. 1\,800\,008 (1--14), Oct. 2018.

\bibitem{paniFullyTextilePEDOT2016}
D.~Pani, A.~Dessi, J.~F. {Saenz-Cogollo}, G.~Barabino, B.~Fraboni, and
  A.~Bonfiglio, ``Fully {{Textile}}, {{PEDOT}}:{{PSS Based Electrodes}} for
  {{Wearable ECG Monitoring Systems}},'' \emph{IEEE Transactions on Biomedical
  Engineering}, vol.~63, no.~3, Mar. 2016.

\bibitem{chlaihawiDevelopmentPrintedFlexible2018}
A.~A. Chlaihawi, B.~B. Narakathu, S.~Emamian, B.~J. Bazuin, and M.~Z. Atashbar,
  ``Development of printed and flexible dry {{ECG}} electrodes,'' \emph{Sensing
  and Bio-Sensing Research}, vol.~20, pp. 9--15, Sep. 2018.

\bibitem{tsukadaValidationWearableTextile2019}
Y.~T. Tsukada, M.~Tokita, H.~Murata, Y.~Hirasawa, K.~Yodogawa, Y.-k. Iwasaki,
  K.~Asai, W.~Shimizu, N.~Kasai, H.~Nakashima, and S.~Tsukada, ``Validation of
  wearable textile electrodes for {{ECG}} monitoring,'' \emph{Heart and
  Vessels}, vol.~34, no.~7, Jul. 2019.

\bibitem{leePUNanowebbasedTextile2019}
E.~Lee and G.~Cho, ``{{PU}} nanoweb-based textile electrode treated with
  single-walled carbon nanotube/silver nanowire and its application to {{ECG}}
  monitoring,'' \emph{Smart Materials and Structures}, vol.~28, no.~4, p.
  045004, Apr. 2019.

\bibitem{tasneemLowPowerOnChipECG2020}
N.~T. Tasneem, S.~A. Pullano, C.~D. Critello, A.~S. Fiorillo, and I.~Mahbub,
  ``A {{Low-Power On-Chip ECG Monitoring System Based}} on {{MWCNT}}/{{PDMS Dry
  Electrodes}},'' \emph{IEEE Sensors Journal}, vol.~20, no.~21, pp.
  12\,799--12\,806, Nov. 2020.

\bibitem{choPerformanceEvaluationTextileBased2011}
G.~Cho, K.~Jeong, M.~J. Paik, Y.~Kwun, and M.~Sung, ``Performance
  {{Evaluation}} of {{Textile-Based Electrodes}} and {{Motion Sensors}} for
  {{Smart Clothing}},'' \emph{IEEE Sensors Journal}, vol.~11, no.~12, p. 3183,
  Dec. 2011.

\bibitem{dirienzoEvaluationTextilebasedWearable2013}
M.~Di~Rienzo, V.~Racca, F.~Rizzo, B.~Bordoni, G.~Parati, P.~Castiglioni,
  P.~Meriggi, and M.~Ferratini, ``Evaluation of a textile-based wearable system
  for the electrocardiogram monitoring in cardiac patients,'' \emph{EP
  Europace}, vol.~15, no.~4, Apr. 2013.

\bibitem{leeFlexibleCapacitiveElectrodes2014}
J.~Lee, J.~Heo, W.~Lee, Y.~Lim, Y.~Kim, and K.~Park, ``Flexible {{Capacitive
  Electrodes}} for {{Minimizing Motion Artifacts}} in {{Ambulatory
  Electrocardiograms}},'' \emph{Sensors}, vol.~14, no.~8, Aug. 2014.

\bibitem{wederEmbroideredElectrodeSilver2015}
M.~Weder, D.~Hegemann, M.~Amberg, M.~Hess, L.~Boesel, R.~Ab{\"a}cherli,
  V.~Meyer, and R.~Rossi, ``Embroidered {{Electrode}} with
  {{Silver}}/{{Titanium Coating}} for {{Long-Term ECG Monitoring}},''
  \emph{Sensors}, vol.~15, no.~1, Jan. 2015.

\bibitem{boehmNovel12LeadECG2016}
A.~Boehm, X.~Yu, W.~Neu, S.~Leonhardt, and D.~Teichmann, ``A {{Novel}}
  12-{{Lead ECG T-Shirt}} with {{Active Electrodes}},'' \emph{Electronics},
  vol.~5, no.~4, p.~75, Nov. 2016.

\bibitem{xiaoPerformanceEvaluationPlain2017}
X.~Xiao, S.~Pirbhulal, K.~Dong, W.~Wu, and X.~Mei, ``Performance {{Evaluation}}
  of {{Plain Weave}} and {{Honeycomb Weave Electrodes}} for {{Human ECG
  Monitoring}},'' \emph{Journal of Sensors}, vol. 2017, 2017.

\bibitem{sunWearableHshirtExercise2017}
F.~Sun, C.~Yi, W.~Li, and Y.~Li, ``A wearable {{H-shirt}} for exercise {{ECG}}
  monitoring and individual lactate threshold computing,'' \emph{Computers in
  Industry}, vol. 92--93, pp. 1--11, Nov. 2017.

\bibitem{anHybridTextileElectrode2018}
X.~An and G.~Stylios, ``A {{Hybrid Textile Electrode}} for
  {{Electrocardiogram}} ({{ECG}}) {{Measurement}} and {{Motion Tracking}},''
  \emph{Materials}, vol.~11, no.~10, p. 1887, Oct. 2018.

\bibitem{achilliDesignCharacterizationScreenPrinted2018}
A.~Achilli, A.~Bonfiglio, and D.~Pani, ``Design and {{Characterization}} of
  {{Screen-Printed Textile Electrodes}} for {{ECG Monitoring}},'' \emph{IEEE
  Sensors Journal}, vol.~18, no.~10, pp. 4097--4107, May 2018.

\bibitem{ankhiliAmbulatoryEvaluationECG2019}
A.~Ankhili, X.~Tao, C.~Cochrane, V.~Koncar, D.~Coulon, and J.-M. Tarlet,
  ``Ambulatory {{Evaluation}} of {{ECG Signals Obtained Using Washable
  Textile-Based Electrodes Made}} with {{Chemically Modified PEDOT}}:{{PSS}},''
  \emph{Sensors}, vol.~19, no.~2, p. 416, Jan. 2019.

\bibitem{liuIntegratedDesignMultiChannel2020}
L.~Liu, X.~Zhu, and Q.~Xia, ``An {{Integrated Design}} of {{Multi-Channel ECG
  Sensor}} on {{Smart Garment}},'' in \emph{Proceedings of the 2020 10th
  {{International Conference}} on {{Biomedical Engineering}} and
  {{Technology}}}.\hskip 1em plus 0.5em minus 0.4em\relax Tokyo Japan: ACM,
  Sep. 2020, pp. 316--320.

\bibitem{arquillaDetectionCompleteECG2021}
K.~Arquilla, L.~Devendorf, A.~K. Webb, and A.~P. Anderson, ``Detection of the
  {{Complete ECG Waveform}} with {{Woven Textile Electrodes}},''
  \emph{Biosensors}, vol.~11, no.~9, p. 331, Sep. 2021.

\bibitem{blasingECGPerformanceSimultaneous2022}
D.~Bl{\"a}sing, A.~Buder, J.~E. Reiser, M.~Nisser, S.~Derlien, and M.~Vollmer,
  ``{{ECG}} performance in simultaneous recordings of five wearable devices
  using a new morphological noise-to-signal index and {{Smith-Waterman-based
  RR}} interval comparisons,'' \emph{PLOS ONE}, vol.~17, no.~10, Oct. 2022.

\bibitem{neriComparisonSingleLeadECG2024}
L.~Neri, I.~Corazza, M.~T. Oberdier, J.~Lago, I.~Gallelli, A.~F. Cicero,
  I.~Diemberger, A.~Orro, A.~Beker, N.~Paolocci, H.~R. Halperin, and C.~Borghi,
  ``Comparison {{Between}} a {{Single-Lead ECG Garment Device}} and a {{Holter
  Monitor}}: {{A Signal Quality Assessment}},'' \emph{Journal of Medical
  Systems}, vol.~48, no.~1, p.~57, May 2024.

\bibitem{polaTextileElectrodesECG2007}
T.~Pola and J.~Vanhala, ``Textile {{Electrodes}} in {{ECG Measurement}},'' in
  \emph{2007 3rd {{International Conference}} on {{Intelligent Sensors}},
  {{Sensor Networks}} and {{Information}}}.\hskip 1em plus 0.5em minus
  0.4em\relax Melbourne, Australia: IEEE, 2007.

\bibitem{qinNovelWearableElectrodes2018}
H.~Qin, J.~Li, B.~He, J.~Sun, L.~Li, and L.~Qian, ``Novel {{Wearable Electrodes
  Based}} on {{Conductive Chitosan Fabrics}} and {{Their Application}} in
  {{Smart Garments}},'' \emph{Materials}, vol.~11, no.~3, p. 370, Mar. 2018.

\bibitem{gunnarssonSeamlesslyIntegratedTextile2023}
E.~Gunnarsson, K.~R{\"o}dby, and F.~Seoane, ``Seamlessly integrated textile
  electrodes and conductive routing in a garment for electrostimulation:
  Design, manufacturing and evaluation,'' \emph{Scientific Reports}, vol.~13,
  no.~1, Oct. 2023.

\bibitem{biharFullyPrintedElectrodes2017}
E.~Bihar, T.~Roberts, E.~Ismailova, M.~Saadaoui, M.~Isik, A.~Sanchez-Sanchez,
  D.~Mecerreyes, T.~Herv{\'e}, J.~B. De~Graaf, and G.~G. Malliaras, ``Fully
  {{Printed Electrodes}} on {{Stretchable Textiles}} for {{Long}}-{{Term
  Electrophysiology}},'' \emph{Advanced Materials Technologies}, vol.~2, no.~4,
  p. 1600251, Apr. 2017.

\bibitem{takeshitaRelationshipContactPressure2019}
T.~Takeshita, M.~Yoshida, Y.~Takei, A.~Ouchi, A.~Hinoki, H.~Uchida, and
  T.~Kobayashi, ``Relationship between {{Contact Pressure}} and {{Motion
  Artifacts}} in {{ECG Measurement}} with {{Electrostatic Flocked Electrodes
  Fabricated}} on {{Textile}},'' \emph{Scientific Reports}, vol.~9, no.~1, Apr.
  2019.

\bibitem{yarnozMoreReasonsWhy2008}
M.~J. Yarnoz and A.~B. Curtis, ``More {{Reasons Why Men}} and {{Women Are Not}}
  the {{Same}} ({{Gender Differences}} in {{Electrophysiology}} and
  {{Arrhythmias}}),'' \emph{The American Journal of Cardiology}, vol. 101,
  no.~9, May 2008.

\bibitem{kittnarSexRelatedDifferences2023}
O.~Kittnar, ``Sex {{Related Differences}} in {{Electrocardiography}},''
  \emph{Physiological Research}, Jul. 2023.

\bibitem{prajapatiSexDifferencesHeart2022}
C.~Prajapati, J.~Koivum{\"a}ki, M.~{Pekkanen-Mattila}, and
  K.~{Aalto-Set{\"a}l{\"a}}, ``Sex differences in heart: From basics to
  clinics,'' \emph{European Journal of Medical Research}, vol.~27, no.~1, Nov.
  2022.

\bibitem{rautaharjuStandardizedProcedureLocating1998}
P.~M. Rautaharju, L.~Park, F.~S. Rautaharju, and R.~Crow, ``A standardized
  procedure for locating and documenting ecg chest electrode positions:
  {{Consideration}} of the effect of breast tissue on ecg amplitudes in
  women,'' \emph{Journal of Electrocardiology}, vol.~31, no.~1, Jan. 1998.

\bibitem{colacoFalsePositiveECG2000}
R.~Colaco, P.~Reay, C.~Beckett, T.~C. Aitchison, and P.~W. Macfarlane, ``False
  positive {{ECG}} reports of anterior myocardial infarction in women,''
  \emph{Journal of Electrocardiology}, vol.~33, Jan. 2000.

\bibitem{drewPitfallsArtifactsElectrocardiography2006}
B.~J. Drew, ``Pitfalls and {{Artifacts}} in {{Electrocardiography}},''
  \emph{Cardiology Clinics}, vol.~24, no.~3, Aug. 2006.

\bibitem{DINISO85591}
``{{DIN EN ISO}} 8559-1:2021-04, {{Gr{\"o}{\ss}enbezeichnung}} von
  {{Bekleidung}}\_- {{Teil}}\_1: {{Anthropometrische Definition}} f{\"u}r
  {{K{\"o}rperma{\ss}e}} ({{ISO}}\_8559-1:2017); {{Deutsche Fassung
  EN}}\_{{ISO}}\_8559-1:2020.''

\bibitem{jaeggiConcurrentValidityNBack2010}
S.~M. Jaeggi, M.~Buschkuehl, W.~J. Perrig, and B.~Meier, ``The concurrent
  validity of the {{N-Back}} task as a working memory measure,'' \emph{Memory},
  vol.~18, no.~4, pp. 394--412, 2010.

\bibitem{cliffordSignalQualityIndices2011}
{\relax GD}.~Clifford, D.~Lopez, Q.~Li, and I.~Rezek, ``Signal quality indices
  and data fusion for determining acceptability of electrocardiograms collected
  in noisy ambulatory environments,'' in \emph{Computing in {{Cardiology}}},
  Hangzhou, China, 2011.

\bibitem{beharSingleChannelECG2012}
J.~Behar, J.~Oster, Q.~Li, and G.~D. Clifford, ``A single channel {{ECG}}
  quality metric,'' in \emph{Computing in {{Cardiology}}}, vol. 2012 Computing
  in Cardiology, Krakow, Poland, 2012, pp. 381--384.

\bibitem{zhaoSQIQualityEvaluation2018}
Z.~Zhao and Y.~Zhang, ``{{SQI Quality Evaluation Mechanism}} of {{Single-Lead
  ECG Signal Based}} on {{Simple Heuristic Fusion}} and {{Fuzzy Comprehensive
  Evaluation}},'' \emph{Frontiers in Physiology}, vol.~9, p. 727, Jun. 2018.

\bibitem{haynQRSDetectionBased2012}
D.~Hayn, B.~Jammerbund, and G.~Schreier, ``{{QRS}} detection based {{ECG}}
  quality assessment,'' \emph{Physiological Measurement}, vol.~33, no.~9, Sep.
  2012.

\bibitem{moodyXQRSAlgorithm2022}
G.~B. Moody, ``The {{XQRS}} algorithm.'' https://wfdb.readthedocs.io/en/
  latest/processing.html, Jun. 2022.

\bibitem{rodriguesLowComplexityRpeakDetection2021}
T.~Rodrigues, S.~Samoutphonh, H.~Silva, and A.~Fred, ``A {{Low-Complexity
  R-peak Detection Algorithm}} with {{Adaptive Thresholding}} for {{Wearable
  Devices}},'' in \emph{2020 25th {{International Conference}} on {{Pattern
  Recognition}} ({{ICPR}})}.\hskip 1em plus 0.5em minus 0.4em\relax Milan,
  Italy: IEEE, Jan. 2021, pp. 1--8.

\bibitem{shafferOverviewHeartRate2017}
F.~Shaffer and J.~P. Ginsberg, ``An {{Overview}} of {{Heart Rate Variability
  Metrics}} and {{Norms}},'' \emph{Frontiers in Public Health}, vol.~5, p. 258,
  Sep. 2017.

\bibitem{cliffordAdvancedMethodsTools2006}
G.~D. Clifford, F.~Azuaje, and P.~McSharry, Eds., \emph{Advanced Methods and
  Tools for {{ECG}} Data Analysis}.\hskip 1em plus 0.5em minus 0.4em\relax
  Boston: Artech House, 2006.

\bibitem{hamiltonOpenSourceECG2002}
P.~Hamilton, ``Open source {{ECG}} analysis,'' in \emph{Computers in
  {{Cardiology}}}.\hskip 1em plus 0.5em minus 0.4em\relax Memphis, TN, USA:
  IEEE, 2002, pp. 101--104.

\bibitem{minkaEstimatingDirichletDistribution2012}
T.~P. Minka, ``Estimating a {{Dirichlet}} distribution,'' 2012.

\bibitem{yangInsightContactImpedance2022}
L.~Yang, L.~Gan, Z.~Zhang, Z.~Zhang, H.~Yang, Y.~Zhang, and J.~Wu, ``Insight
  into the {{Contact Impedance}} between the {{Electrode}} and the {{Skin
  Surface}} for {{Electrophysical Recordings}},'' \emph{ACS Omega}, vol.~7,
  no.~16, pp. 13\,906--13\,912, Apr. 2022.

\bibitem{goyalDependenceSkinElectrodeContact2022}
K.~Goyal, D.~A. Borkholder, and S.~W. Day, ``Dependence of {{Skin-Electrode
  Contact Impedance}} on {{Material}} and {{Skin Hydration}},'' \emph{Sensors},
  vol.~22, no.~21, p. 8510, Nov. 2022.

\bibitem{toosiFalseSTElevation2008}
M.~S. Toosi and M.~T. Sochanski, ``False {{ST}} elevation in a modified 12-lead
  surface electrocardiogram,'' \emph{Journal of Electrocardiology}, vol.~41,
  no.~3, pp. 197--201, May 2008.

\bibitem{salviComparison5Methods2011}
V.~Salvi, D.~R. Karnad, G.~K. Panicker, M.~Natekar, P.~Hingorani, V.~Kerkar,
  A.~Ramasamy, M.~De~Vries, T.~Zumbrunnen, S.~Kothari, and D.~Narula,
  ``Comparison of 5 methods of {{QT}} interval measurements on
  electrocardiograms from a thorough {{QT}}/{{QTc}} study: Effect on assay
  sensitivity and categorical outliers,'' \emph{Journal of Electrocardiology},
  vol.~44, no.~2, pp. 96--104, Mar. 2011.

\bibitem{liSolvingInverseProblem2025}
L.~Li, J.~Camps, B.~Rodriguez, and V.~Grau, ``Solving the {{Inverse Problem}}
  of {{Electrocardiography}} for {{Cardiac Digital Twins}}: {{A Survey}},''
  \emph{IEEE Reviews in Biomedical Engineering}, vol.~18, pp. 316--336, 2025.

\bibitem{marsanovaBrnoUniversityTechnology2021}
L.~Mar{\v s}{\'a}nov{\'a}, A.~Nemcova, R.~Smisek, L.~Smital, and M.~Vitek,
  ``Brno {{University}} of {{Technology ECG Signal Database}} with
  {{Annotations}} of {{P Wave}} ({{BUT PDB}}),'' 2021.

\end{thebibliography}


\begin{thebibliography}{10}
\providecommand{\url}[1]{#1}
\csname url@samestyle\endcsname
\providecommand{\newblock}{\relax}
\providecommand{\bibinfo}[2]{#2}
\providecommand{\BIBentrySTDinterwordspacing}{\spaceskip=0pt\relax}
\providecommand{\BIBentryALTinterwordstretchfactor}{4}
\providecommand{\BIBentryALTinterwordspacing}{\spaceskip=\fontdimen2\font plus
\BIBentryALTinterwordstretchfactor\fontdimen3\font minus
  \fontdimen4\font\relax}
\providecommand{\BIBforeignlanguage}[2]{{%
\expandafter\ifx\csname l@#1\endcsname\relax
\typeout{** WARNING: IEEEtran.bst: No hyphenation pattern has been}%
\typeout{** loaded for the language `#1'. Using the pattern for}%
\typeout{** the default language instead.}%
\else
\language=\csname l@#1\endcsname
\fi
#2}}
\providecommand{\BIBdecl}{\relax}
\BIBdecl

\bibitem{cliffordAdvancedMethodsTools2006}
G.~D. Clifford, F.~Azuaje, and P.~McSharry, Eds., \emph{Advanced Methods and
  Tools for {{ECG}} Data Analysis}.\hskip 1em plus 0.5em minus 0.4em\relax
  Boston: Artech House, 2006.

\bibitem{satijaReviewSignalProcessing2018}
U.~Satija, B.~Ramkumar, and M.~S. Manikandan, ``A {{Review}} of {{Signal
  Processing Techniques}} for {{Electrocardiogram Signal Quality
  Assessment}},'' \emph{IEEE Reviews in Biomedical Engineering}, vol.~11, 2018.

\bibitem{luoReviewElectrocardiogramFiltering2010}
S.~Luo and P.~Johnston, ``A review of electrocardiogram filtering,''
  \emph{Journal of Electrocardiology}, vol.~43, no.~6, Nov. 2010.

\bibitem{masonRecommendationsStandardizationInterpretation2007}
J.~W. Mason, E.~W. Hancock, and L.~S. Gettes, ``Recommendations for the
  {{Standardization}} and {{Interpretation}} of the {{Electrocardiogram}}:
  {{Part II}}: {{Electrocardiography Diagnostic Statement List}}: {{A
  Scientific Statement From}} the {{American Heart Association
  Electrocardiography}} and {{Arrhythmias Committee}}, {{Council}} on
  {{Clinical Cardiology}}; the {{American College}} of {{Cardiology
  Foundation}}; and the {{Heart Rhythm Society}}: {{{\emph{Endorsed}}}}{\emph{
  by the }}{{{\emph{International Society}}}}{\emph{ for
  }}{{{\emph{Computerized Electrocardiology}}}},'' \emph{Circulation}, vol.
  115, no.~10, Mar. 2007.

\bibitem{addisonWaveletTransformsECG2005}
P.~S. Addison, ``Wavelet transforms and the {{ECG}}: A review,''
  \emph{Physiological Measurement}, vol.~26, no.~5, Oct. 2005.

\bibitem{liDeScoDECGDeepScoreBased2024}
H.~Li, G.~Ditzler, J.~Roveda, and A.~Li, ``{{DeScoD-ECG}}: {{Deep Score-Based
  Diffusion Model}} for {{ECG Baseline Wander}} and {{Noise Removal}},''
  \emph{IEEE Journal of Biomedical and Health Informatics}, 2024.

\bibitem{oppeltCombiningScatterTransform2020}
M.~P. Oppelt, M.~Riehl, F.~P. Kemeth, and J.~Steffan, ``Combining {{Scatter
  Transform}} and {{Deep Neural Networks}} for {{Multilabel Electrocardiogram
  Signal Classification}},'' in \emph{2020 {{Computing}} in
  {{Cardiology}}}.\hskip 1em plus 0.5em minus 0.4em\relax Rimini, Italy: IEEE,
  2020.

\bibitem{panRealTimeQRSDetection1985}
J.~Pan and W.~J. Tompkins, ``A {{Real-Time QRS Detection Algorithm}},''
  \emph{IEEE Transactions on Biomedical Engineering}, vol. BME-32, no.~3, pp.
  230--236, Mar. 1985.

\bibitem{moodyMITBIHArrhythmiaDatabase1992}
G.~B. Moody and R.~G. Mark, ``{{MIT-BIH Arrhythmia Database}},'' 1992.

\bibitem{hamiltonOpenSourceECG2002}
P.~Hamilton, ``Open source {{ECG}} analysis,'' in \emph{Computers in
  {{Cardiology}}}.\hskip 1em plus 0.5em minus 0.4em\relax Memphis, TN, USA:
  IEEE, 2002, pp. 101--104.

\bibitem{rodriguesLowComplexityRpeakDetection2021}
T.~Rodrigues, S.~Samoutphonh, H.~Silva, and A.~Fred, ``A {{Low-Complexity
  R-peak Detection Algorithm}} with {{Adaptive Thresholding}} for {{Wearable
  Devices}},'' in \emph{2020 25th {{International Conference}} on {{Pattern
  Recognition}} ({{ICPR}})}.\hskip 1em plus 0.5em minus 0.4em\relax Milan,
  Italy: IEEE, Jan. 2021, pp. 1--8.

\bibitem{emrichPhysiologyInformedECGDelineation2024}
J.~Emrich, A.~Gargano, T.~Koka, and M.~Muma, ``Physiology-{{Informed ECG
  Delineation Based}} on {{Peak Prominence}},'' in \emph{2024 32nd {{European
  Signal Processing Conference}} ({{EUSIPCO}})}.\hskip 1em plus 0.5em minus
  0.4em\relax Lyon, France: IEEE, Aug. 2024, pp. 1402--1406.

\bibitem{makowskiNeuroKit2PythonToolbox2021}
D.~Makowski, T.~Pham, Z.~J. Lau, J.~C. Brammer, F.~Lespinasse, H.~Pham,
  C.~Sch{\"o}lzel, and S.~H.~A. Chen, ``{{NeuroKit2}}: {{A Python}} toolbox for
  neurophysiological signal processing,'' \emph{Behavior Research Methods},
  vol.~53, no.~4, pp. 1689--1696, Aug. 2021.

\bibitem{minkaEstimatingDirichletDistribution2012}
T.~P. Minka, ``Estimating a {{Dirichlet}} distribution,'' 2012.

\bibitem{watanabeTreeStructuredParzenEstimator2023}
S.~Watanabe, ``Tree-{{Structured Parzen Estimator}}: {{Understanding Its
  Algorithm Components}} and {{Their Roles}} for {{Better Empirical
  Performance}},'' May 2023.

\end{thebibliography}

\end{document}

\title{Supplementary Materials}
\author{}

\maketitle

\section{Signal Preprocessing}
This work does not primarily focus on improving, filtering, or enhancing signal quality by the development of
novel filtering techniques or fiducial point extraction algorithms. Instead, we rely on established state-of-the-art
signal processing methods for \gls{ecg} signals.

\begin{figure}[ht!]
    \centering
    \includegraphics[]{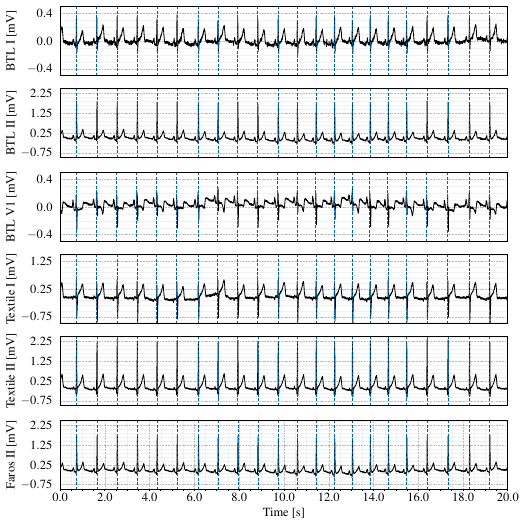}
    \caption{Visualization of recorded channels and devices. The signal shows different signal amplitude values during
a sitting phase with a resting heart rate and annotated peak locations.}
    \label{fig:raw_ecg_detection_results}
\end{figure}

\subsection{Signal Quality and Filtering}
Noise in \gls{ecg} signals manifests on different time scales with characteristic spectral structures
\cite{cliffordAdvancedMethodsTools2006, satijaReviewSignalProcessing2018}. Baseline wander artifacts are caused
by changes in skin-electrode impedance e.g. by sweating and other types of slow drifts, that are caused
by body movements or respiration. These artifacts are typically present in the spectral range between $0.01$ and
$1$ Hz \cite{luoReviewElectrocardiogramFiltering2010}. It is noteworthy that artifacts
below a threshold of $0.67$ Hz can be removed with linear digital zero-phase distortion filters according
to the \gls{ansi_aami} recommendations for standard clinical \gls{ecg}s \cite{masonRecommendationsStandardizationInterpretation2007}. 
However, spectral components above this threshold or filters with a higher cutoff frequency can distort
ST-segment related abnormalities.
Another common artifact is caused by power line interference. This artifact manifests itself in a
narrowband spectral component at $50$ Hz or $60$ Hz and its harmonics and is especially critical as
it can distort morphological signal properties such as the P-wave. These artifacts are commonly
filtered with notch filters \cite{luoReviewElectrocardiogramFiltering2010}. Artifacts
caused by the electrical activity of skeletal muscles are prominently present in
the spectral bandwidth between $20$ Hz and $10$ kHz and can therefore interfere with ECG signal information
\cite{cliffordAdvancedMethodsTools2006}. The research community has developed different methods to mitigate the effect
of these artifacts, like removing them with adaptive filters or wavelet transformations
\cite{addisonWaveletTransformsECG2005}, consider practical aspects like changing the electrode positions for certain
activities, or utilizing end-to-end trained deep learning models \cite{liDeScoDECGDeepScoreBased2024,%
oppeltCombiningScatterTransform2020}, that either denoise the \gls{ecg} for further processing or learning embeddings
that are not prone to noise.

We opted to only filter baseline noise using a forward and backward highpass filter with a cutoff frequency of $0.67$
Hz and recorded the signals in our lab apart from potential $50$Hz (the study was conducted in Germany) noise sources.
The precaution of powerline noise filtering was not necessary in our setup, as the study was conducted away from power
lines and no phases exhibited visible powerline noise. However, such noise could potentially be introduced during the
\emph{Walking} or \emph{Running} activities on the treadmill.

\section{Supplementary Results}
\subsection{Signal Quality Indices}
\begin{figure}[ht!]
    \centering
    \includegraphics[]{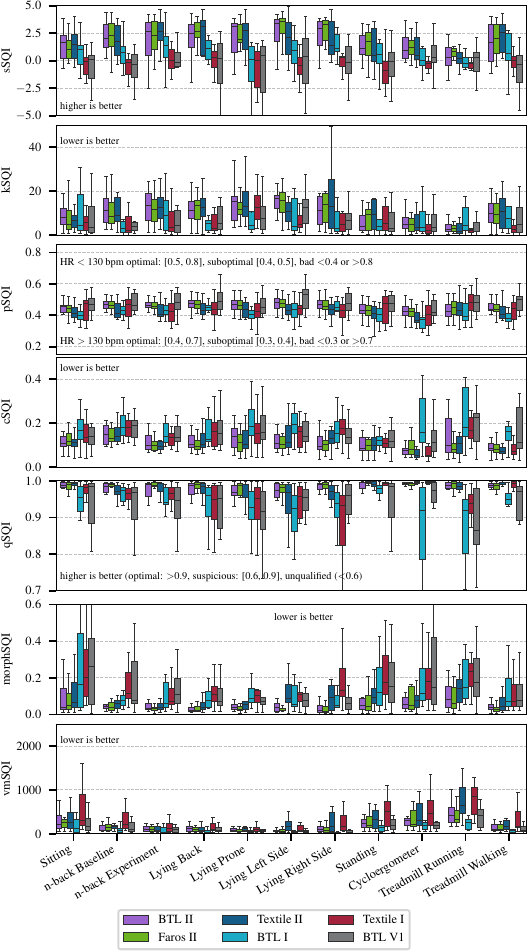}
    \caption{\gls{sqis} for female subjects.}
    \label{fig:female-sqis}
\end{figure}

\begin{figure}[ht!]
    \centering
    \includegraphics[]{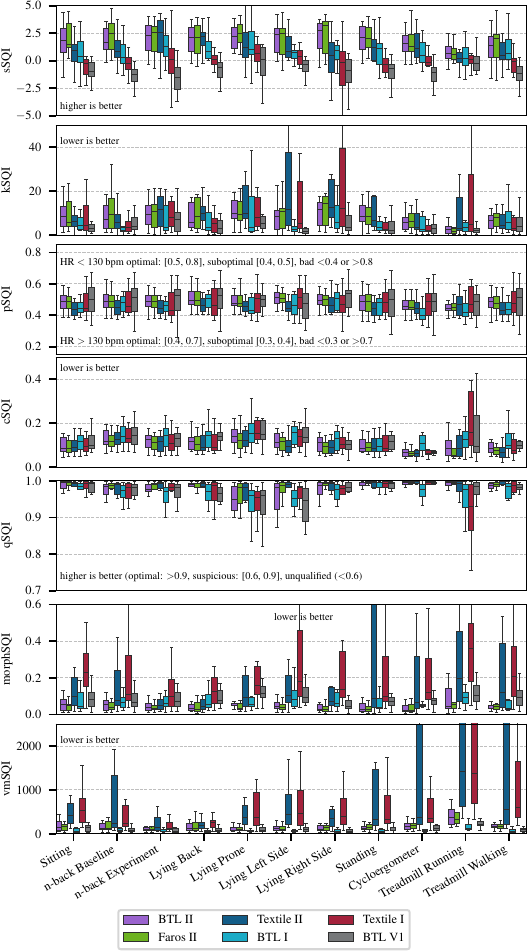}
    \caption{\gls{sqis} for male subjects.}
    \label{fig:male-sqis}
\end{figure}

To complement the findings presented in the main article, we provide the \gls{sqis} for female and male subjects
separately in Sup.~Figures~\ref{fig:female-sqis} and~\ref{fig:male-sqis}, respectively.
Across all systems—including the reference systems BTL and Faros, as well as the textile wearable garment—the general
trend persists: lead~II demonstrates superior performance for these signal quality metrics.
Notably, within the male subgroup, the textile garment exhibits significantly lower morphology-based \gls{sqis} in
both lead~I and Lead~II compared to the female subgroup. This observation warrants further investigation, as factors
such as garment fit or body shape may contribute to the observed reduction in signal quality indices. Future studies
should address these potential influences to better understand the underlying causes of this performance disparity.

\subsection{R-Peak detection and Deliniation}
R-peak detection in \gls{ecg} signals is essential for accurate heart rate analysis and the diagnosis of cardiac
abnormalities, as it identifies the QRS complex peaks associated with ventricular depolarization. The Pan-Tompkins
algorithm remains a cornerstone in this field, employing a sequence of filtering, differentiation, and adaptive
thresholding to achieve reliable real-time R-peak detection \cite{panRealTimeQRSDetection1985}. Building upon this
foundation, advanced approaches such as wavelet transforms and machine learning techniques have been developed to
further enhance detection robustness in the presence of noise and signal variability. The widespread adoption and
validation of these algorithms have been facilitated by benchmark datasets, most notably the MIT-BIH arrhythmia
database \cite{moodyMITBIHArrhythmiaDatabase1992}, which continues to support progress and standardization in
R-peak detection research.
Nowadays, several publicly available software packages provide a variety of state-of-the-art R-peak detection
algorithms, which often differ only in minor methodological improvements. Performance comparisons reveal that no single
algorithm consistently outperforms others across all datasets; rather, the relative effectiveness of an algorithm can
vary depending on the specific characteristics of the data.

To address potential biases arising from algorithm selection, we adopt the following strategies.
First, for evaluating R-peak detection quality, we employ multiple algorithms specifically, the Hamilton
\cite{hamiltonOpenSourceECG2002}, Pan-Tompkins \cite{panRealTimeQRSDetection1985}, and Rodrigues
\cite{rodriguesLowComplexityRpeakDetection2021} methods and concatenate the detection results to mitigate the
influence of any single algorithm's performance. Second, for our machine learning experiments, we utilize the
Rodrigues algorithm that shows promising results other wearable devices \cite{rodriguesLowComplexityRpeakDetection2021}.
Third, for ECG delineation (fiducial point detection), we follow the approach described in
\cite{emrichPhysiologyInformedECGDelineation2024}. All algorithms are implemented in Python using the NeuroKit2
toolbox \cite{makowskiNeuroKit2PythonToolbox2021}.
In this study, we did not evaluate the performance of the peak detection and delineation algorithms. Instead, we
selected an algorithm previously employed for this specific application and deferred the optimization of detection
algorithms to future work.
\subsection{Computing the median complex}
\begin{figure}[ht!]
    \centering
    \begin{subfigure}[t]{0.49\linewidth}
        \includegraphics[width=\textwidth]{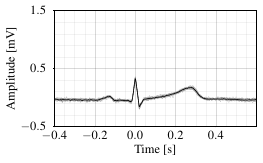}
        \caption{\small BTL Channel I}
        \label{sfig:median-btl-i}
    \end{subfigure}
    \begin{subfigure}[t]{0.49\linewidth}
        \includegraphics[width=\textwidth]{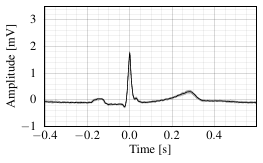}
        \caption{\small BTL Channel II}
        \label{sfig:median-btl-ii}
    \end{subfigure}
    \begin{subfigure}[t]{0.49\linewidth}
        \includegraphics[width=\textwidth]{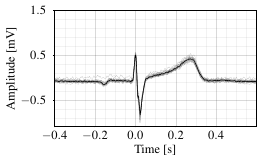}
        \caption{\small Textile Channel I}
        \label{sfig:median-cardiotextil-i}
    \end{subfigure}
    \begin{subfigure}[t]{0.49\linewidth}
        \includegraphics[width=\textwidth]{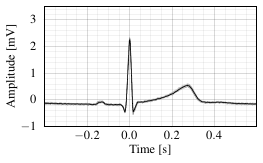}
        \caption{\small Textile Channel II}
        \label{sfig:median-cardiotextil-ii}
    \end{subfigure}
\caption{Heart beats aligned on the R-peaks and computed median complex for a $30\,\mathrm{s}$ window during rest of
the same subject.}
\label{fig:medians}
\end{figure}

\begin{figure}[ht!]
    \centering
    \begin{subfigure}[t]{0.49\linewidth}
        \includegraphics[width=\textwidth]{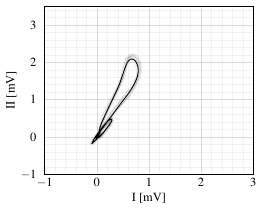}
        \caption{\small BTL I/II}
        \label{sfig:vector-btl-i}
    \end{subfigure}
    \begin{subfigure}[t]{0.49\linewidth}
        \includegraphics[width=\textwidth]{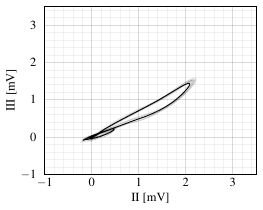}
        \caption{\small BTL II/III}
        \label{sfig:vector-btl-ii}
    \end{subfigure}
    \begin{subfigure}[t]{0.49\linewidth}
        \includegraphics[width=\textwidth]{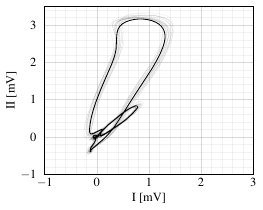}
        \caption{\small Textile I/II}
        \label{sfig:vector-cardiotextil-i}
    \end{subfigure}
    \begin{subfigure}[t]{0.49\linewidth}
        \includegraphics[width=\textwidth]{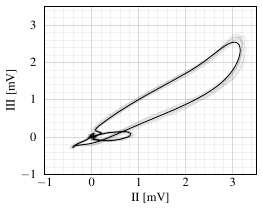}
        \caption{\small Textile II/III}
        \label{sfig:vector-cardiotextil-ii}
    \end{subfigure}
\caption{Vector visualizations over complexes within a $30\,\mathrm{s}$ window and median complex during rest of
the same subject.}
\label{fig:medians_vector}
\end{figure}

Throughout this paper, we employ the computation of a median complex, for example, when performing Procrustes
alignment between two \gls{ecg} devices. We compute these median complexes by first extracting the R-peak locations
and which in turn are then used to align the heart beats, as illustrated in gray in Sup.~Fig.~\ref{fig:medians}. The
median complex is subsequently computed over all aligned complexes.
These median complexes can then be shown in the I/II planes (see Sup.~Fig.~\ref{fig:medians_vector}) and rotated
such that they better align with each other.

\begin{figure}[ht!]
    \centering
    \includegraphics[width=\linewidth]{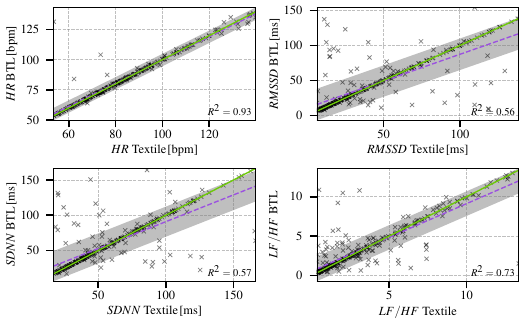}
    \caption{Correlation of physiological parameters measured in Lead II of two devices, the reference system
BTL on the y-axis and the textile on the x-axis. With optimal correlation (green) and actual fit (purple).}
    \label{fig:correlations_physiological}
\end{figure}
\subsection{Recording data}
The sampling frequency of the \gls{ecg} devices employed in this study varied among the platforms utilized.
Specifically, the BTL device operated at a sampling frequency of $500\,\mathrm{Hz}$, the Faros device at
$1000\,\mathrm{Hz}$, and the textile device at $400\,\mathrm{Hz}$. To ensure methodological consistency and facilitate
subsequent data analyses, all \gls{ecg} recordings were resampled to a uniform frequency of $1000\,\mathrm{Hz}$. This
standardization was chosen for simplicity, as computational constraints associated with higher sampling rates were not
a limiting factor in the context of this post-hoc analysis.
\subsection{Synchronization}
The BTL device was consistently initiated first and terminated last during each recording session. Subsequently, either
the Faros or the textile were started and stopped independently. To achieve temporal alignment among recordings,
data from the Faros and textile were divided into $30\,\mathrm{s}$ segments, which were then convolved with the raw BTL
signal. Given the inherent irregularity of heart rate, these segments intersect with the BTL signal only at specific
points, producing a pronounced maximum in the correlation function. The temporal shift corresponding to this maximum
was used to synchronize the signals across devices.
The accuracy of this alignment method was validated in several ways. A button on the BTL system was pressed after
the initiation of the other devices and the start of the study protocol, enabling the extraction of an event
marker for subsequent comparison.
Additional confirmation was obtained through visual inspection and by referencing the onboard clocks of the
devices.
Notably, this approach eliminates the requirement for any physical electrical connections or wiring between devices.
Furthermore, over the relatively short duration of our recordings, we did not observe any significant clock drift among
the systems.
\subsection{Peak Detection}
To complement the findings presented in the main document, we provide the raw data of hit, miss, and false alarm rates
in Sup.~Table~\ref{tab:sex_peak_performance_btl_ecg_ii_cardiotextil_ecg_ii} and
Sup.~Table~\ref{tab:sex_peak_performance_btl_ecg_ii_faros_ecg_ii}, reporting both mean and standard deviation values.
The results indicate a marked decline in the hit rates and an increase in miss and false alarm rates for the
BTL/textile system compared to the BTL/Faros device, particularly during the \emph{Running} and \emph{Lying} phases. In
contrast, performance during the \emph{Sitting} phase and less intensive upright activities, such as the
\emph{Cycloergometer} and \emph{Walking} phases, remains comparable between the two systems.

A comparison of detection performance between female and male participants in our R-peak detection experiments
revealed no significant differences. This finding aligns with our main article, suggesting that for tasks relying
primarily on R-peak detection, such as the assessment of heart rhythm or heart rate variability, sex does not
constitute a major influencing factor.

\begin{table}[!ht]
\renewcommand{\arraystretch}{1.5}
\caption{R-Peak detection performance between BTL lead~II and textile lead~II for both sexs.}
\label{tab:sex_peak_performance_btl_ecg_ii_cardiotextil_ecg_ii}
    \centering
    \begin{NiceTabular}{l|l|CCC}
    \toprule
    ~ & ~ & Hit & Miss & False Alarm \\
    \midrule
\multirow{11}{*}{Female} & Sitting Relaxed & $0.97\pm0.06$ & $0.02\pm0.03$ & $0.02\pm0.03$ \\ 
~ &  Sitting Video & $0.95\pm0.08$ & $0.02\pm0.04$ & $0.03\pm0.06$ \\ 
~ &  $n$-Back & $0.96\pm0.08$ & $0.02\pm0.04$ & $0.02\pm0.05$ \\ 
~ &  Lying Back & $0.95\pm0.08$ & $0.02\pm0.04$ & $0.03\pm0.05$ \\ 
~ &  Lying Left & $0.93\pm0.13$ & $0.03\pm0.05$ & $0.04\pm0.09$ \\ 
~ &  Lying Right & $0.93\pm0.11$ & $0.02\pm0.04$ & $0.05\pm0.09$ \\ 
~ &  Lying Stomach & $0.94\pm0.08$ & $0.03\pm0.05$ & $0.03\pm0.05$ \\ 
~ &  Standing & $0.96\pm0.06$ & $0.02\pm0.05$ & $0.01\pm0.02$ \\ 
~ &  Ergometer & $0.98\pm0.03$ & $0.01\pm0.02$ & $0.01\pm0.01$ \\ 
~ &  Running & $0.84\pm0.26$ & $0.09\pm0.17$ & $0.07\pm0.10$ \\ 
~ &  Walking & $0.96\pm0.09$ & $0.02\pm0.04$ & $0.02\pm0.05$ \\ 
    \midrule
\multirow{11}{*}{Male} & Sitting Relaxed & $0.97\pm0.05$ & $0.02\pm0.04$ & $0.01\pm0.02$ \\ 
~ &  Sitting Video & $0.95\pm0.08$ & $0.02\pm0.04$ & $0.03\pm0.06$ \\ 
~ &  $n$-Back & $0.96\pm0.08$ & $0.02\pm0.04$ & $0.02\pm0.06$ \\ 
~ &  Lying Back & $0.96\pm0.07$ & $0.02\pm0.05$ & $0.02\pm0.04$ \\ 
~ &  Lying Left & $0.93\pm0.13$ & $0.04\pm0.07$ & $0.03\pm0.07$ \\ 
~ &  Lying Right & $0.89\pm0.24$ & $0.05\pm0.09$ & $0.07\pm0.17$ \\ 
~ &  Lying Stomach & $0.88\pm0.18$ & $0.06\pm0.08$ & $0.07\pm0.13$ \\ 
~ &  Standing & $0.96\pm0.07$ & $0.02\pm0.05$ & $0.01\pm0.03$ \\ 
~ &  Ergometer & $0.94\pm0.15$ & $0.04\pm0.10$ & $0.02\pm0.05$ \\ 
~ &  Running & $0.91\pm0.14$ & $0.04\pm0.07$ & $0.05\pm0.07$ \\ 
~ &  Walking & $0.93\pm0.15$ & $0.03\pm0.05$ & $0.04\pm0.12$ \\ 
\bottomrule
    \end{NiceTabular}
\end{table}
\begin{table}[!ht]
\renewcommand{\arraystretch}{1.5}
\caption{R-Peak detection performance between BTL Lead~II and Faros Lead~II for both sexs.}
\label{tab:sex_peak_performance_btl_ecg_ii_faros_ecg_ii}
    \centering
    \begin{NiceTabular}{l|l|CCC}
    \toprule
    ~ & ~ & Hit & Miss & False Alarm \\
    \midrule
\multirow{11}{*}{Female} & Sitting Relaxed & $0.96\pm0.06$ & $0.02\pm0.03$ & $0.02\pm0.04$ \\ 
~ &  Sitting Video & $0.96\pm0.06$ & $0.02\pm0.04$ & $0.02\pm0.03$ \\ 
~ &  $n$-Back & $0.96\pm0.07$ & $0.02\pm0.05$ & $0.01\pm0.03$ \\ 
~ &  Lying Back & $0.96\pm0.06$ & $0.02\pm0.04$ & $0.02\pm0.02$ \\ 
~ &  Lying Left & $0.95\pm0.07$ & $0.03\pm0.05$ & $0.02\pm0.03$ \\ 
~ &  Lying Right & $0.97\pm0.04$ & $0.01\pm0.02$ & $0.01\pm0.02$ \\ 
~ &  Lying Stomach & $0.95\pm0.07$ & $0.02\pm0.04$ & $0.03\pm0.06$ \\ 
~ &  Standing & $0.98\pm0.03$ & $0.01\pm0.02$ & $0.01\pm0.01$ \\ 
~ &  Ergometer & $0.97\pm0.03$ & $0.01\pm0.02$ & $0.01\pm0.02$ \\ 
~ &  Running & $0.91\pm0.17$ & $0.04\pm0.09$ & $0.05\pm0.09$ \\ 
~ &  Walking & $0.97\pm0.05$ & $0.02\pm0.03$ & $0.02\pm0.03$ \\ 
    \midrule
\multirow{11}{*}{Male} & Sitting Relaxed & $0.98\pm0.04$ & $0.01\pm0.02$ & $0.01\pm0.02$ \\ 
~ &  Sitting Video & $0.96\pm0.07$ & $0.02\pm0.04$ & $0.02\pm0.03$ \\ 
~ &  $n$-Back & $0.96\pm0.10$ & $0.02\pm0.05$ & $0.02\pm0.06$ \\ 
~ &  Lying Back & $0.97\pm0.06$ & $0.02\pm0.04$ & $0.02\pm0.04$ \\ 
~ &  Lying Left & $0.95\pm0.10$ & $0.02\pm0.06$ & $0.02\pm0.06$ \\ 
~ &  Lying Right & $0.96\pm0.11$ & $0.03\pm0.06$ & $0.02\pm0.05$ \\ 
~ &  Lying Stomach & $0.93\pm0.12$ & $0.04\pm0.07$ & $0.03\pm0.06$ \\ 
~ &  Standing & $0.98\pm0.03$ & $0.01\pm0.01$ & $0.01\pm0.02$ \\ 
~ &  Ergometer & $0.99\pm0.02$ & $0.01\pm0.01$ & $0.01\pm0.01$ \\ 
~ &  Running & $0.97\pm0.04$ & $0.01\pm0.01$ & $0.02\pm0.02$ \\ 
~ &  Walking & $0.97\pm0.04$ & $0.02\pm0.03$ & $0.01\pm0.02$ \\ 
\bottomrule
    \end{NiceTabular}
\end{table}
As supplementary comparision between our male and female population Table~\ref{tab:male_confusion} and
Table~\ref{tab:female_confusion} are presented in accordance to our main article. 
\begin{table}[!ht]
  \renewcommand{\arraystretch}{1.5}
  \caption{Male R-Peak Detection Matrix}
  \label{tab:male_confusion}
      \centering
      \begin{NiceTabular}{ll|C|C|ll}
      \toprule
~ & ~ & \Block{1-2}{BTL} & ~ & ~ \\ 
~ & ~ & detected & not det. & ~ & ~  \\ 
\midrule
\parbox[t]{2mm}{\multirow{4}{*}{\rotatebox[origin=c]{90}{Textile}}} & detected & $0.929 \pm 0.078$ & $0.002 \pm 0.003$ & \multirow{2}{*}{detected} & \parbox[t]{2mm}{\multirow{4}{*}{\rotatebox[origin=c]{90}{Faros}}} \\ 
\cline{2-4} ~ & not det. & $0.001 \pm 0.002$ & $0.032 \pm 0.049$  & ~ & ~ \\
\cline{2-5} ~ & detected & $0.024 \pm 0.035$ & $0.006 \pm 0.003$ & \multirow{2}{*}{not det.} & ~ \\ 
\cline{2-4} ~ & not det. & $0.006 \pm 0.003$ & - & ~ & ~ \\ 
\bottomrule
    \end{NiceTabular}
\end{table}
\begin{table}[!ht]
  \renewcommand{\arraystretch}{1.5}
  \caption{Female R-Peak Detection Matrix}
  \label{tab:female_confusion}
      \centering
      \begin{NiceTabular}{ll|C|C|ll}
      \toprule
~ & ~ & \Block{1-2}{BTL} & ~ & ~ \\ 
~ & ~ & detected & not det. & ~ & ~  \\ 
\midrule
\parbox[t]{2mm}{\multirow{4}{*}{\rotatebox[origin=c]{90}{Textile}}} & detected & $0.953 \pm 0.039$ & $0.007 \pm 0.014$ & \multirow{2}{*}{detected} & \parbox[t]{2mm}{\multirow{4}{*}{\rotatebox[origin=c]{90}{Faros}}} \\ 
\cline{2-4} ~ & not det. & $0.003 \pm 0.003$ & $0.012 \pm 0.01$  & ~ & ~ \\
\cline{2-5} ~ & detected & $0.006 \pm 0.005$ & $0.008 \pm 0.005$ & \multirow{2}{*}{not det.} & ~ \\ 
\cline{2-4} ~ & not det. & $0.012 \pm 0.014$ & - & ~ & ~ \\ 
\bottomrule
    \end{NiceTabular}
\end{table}
\subsection{Alignment}
\gls{ecg}s acquired using different measurement systems, differing in aspects such as skin-electrode contact quality,
electrode placement, or lead orientation, may exhibit signal variability, even though both capture the same underlying
cardiac electrical activity. This variability can be attributed to factors including tissue composition at the
measurement sites, proximity of the electrodes to the heart, and the electrical properties of the skin-to-electrode
interface.
We therefore can only approximate the transformation between the two recordings as \gls{ecg}s inherently resemble an
inverse problem, as the signals recorded on the skin surface represent the aggregated activity of numerous cardiac
cells, rather than direct measurements of the underlying sources of activation.

Despite these differences, we opted to approximate the underlying signals for comparative analysis, by adopting a
two-dimensional transformation on the leads I/II plane from one system to the other, with lead III being defined
by lead III = lead II - lead I.

To align the recordings from the two systems, we employ a rotational Procrustes
alignment, constraining the rotation matrix according to established mathematical properties: the transpose of the
rotation matrix equals its inverse ($R^T = R^{-1}$), and the determinant of the matrix is unity ($|R| = 1$).
Formally, given an \gls{ecg} recording $\mathbf{A} \in \mathbb{R}^{m \times n}$ and a reference \gls{ecg} 
$\mathbf{B} \in \mathbb{R}^{m \times n}$ where $n = 2$ represents the two approximated leads, and $m$ denotes the
number of samples within a single median complex.

We seek a rotation matrix $\mathbf{R} \in \mathbb{R}^{n \times n}$ that minimizes the Frobenius norm of the difference
between the rotated recording and the reference. This optimization problem is described in Equation~1 in the main 
article.
To further account for gain discrepancies in the recordings attributable to factors like electrode properties, we
introduce a linear scaling such that the maximum magnitude in I/II of $A$ is the same as the maximum magnitude in
lead~I/II of $B$.
\begin{algorithm}[H]
 \caption{Algorithm for median complex projection alignment of textile median complex $\mathbf{A}$ and reference median 
 complex $\mathbf{B}$ using rotation matrix $\mathbf{R}$ and scaling $s$.} 
 \begin{algorithmic}[1]
 \renewcommand{\algorithmicrequire}{\textbf{Input:}}
 \renewcommand{\algorithmicensure}{\textbf{Output:}}
 \REQUIRE $\mathbf{A}$, $\mathbf{B}$
 \ENSURE  $\mathbf{R}$, $\mathbf{s}$
 \STATE $|\mathbf{v}|_\mathbf{A} = \sqrt{I_{A}^2 + II_{A}^2}$
 \STATE $|\mathbf{v}|_\mathbf{B} = \sqrt{I_{B}^2 + II_{B}^2}$
 \STATE $s = \frac{\mathrm{abs}(\mathrm{max}(|\mathbf{v}|_\mathbf{A}))}{\mathrm{abs}(\mathrm{max}(|\mathbf{v}|_\mathbf{A}))}$
 \STATE $\mathbf{A'} = \mathbf{s}\mathbf{A}$
 \STATE $\mathbf{R} \leftarrow	\min(\|\mathbf{A'}\mathbf{R} - \mathbf{B}\|_{F}^2)$
 \RETURN $\mathbf{R}$, $s$
 \end{algorithmic}
 \label{alg:aligment}
\end{algorithm}
As this approximation is to our knowledge not used in literature, we report the quality of the fit in the main paper
using three different metrics: First, the squared Frobenius norm error of $N$ samples and $D$ dimension lead vector of
the difference matrix is computed as
\begin{equation}
    \| A - B \|_F^2 = \sum_{i=1}^N \sum_{j=1}^D (a_{ij} - b_{ij})^2
\end{equation}
providing a global measure of the aggregate deviation across all leads and time points. Second, the root-mean-square
deviation (RMSD) is calculated to offer a normalized metric of average discrepancy, defined a
\begin{equation}
    \mathrm{RMSD} = \sqrt{\frac{1}{N} \sum_{i=1}^N \sum_{j=1}^D (a_{ij} - b_{ij})^2}
\end{equation}
Lastly, we utilize the cosine similarity, $S_C$, to quantify the degree of alignment in signal directionality between
the two recordings. For $N$-dimensional ECG signal vectors $A$ and $B$, this is computed as
\begin{equation}
    S_C = \frac{1}{N} \sum_{i=1}^N \frac{\sum_{j=1}^D a_{ij} b_{ij}}{\sqrt{\sum_{j=1}^D a_{ij}^2} \sqrt{\sum_{j=1}^D b_{ij}^2}}
\end{equation}
\begin{figure}[ht!]
\centering
  \includegraphics{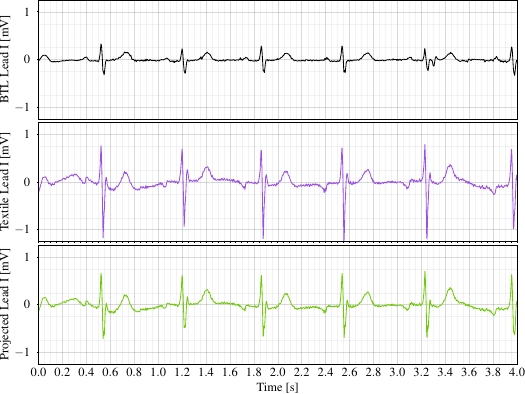}
  \caption{BTL and Textile lead I and the projected version after rotating and scaling.}
  \label{afig:proj}
\end{figure}
We present lead I of the rotated and scaled textile recording alongside the reference system in Sup.~Fig.~\ref{afig:proj}.
The results indicate that, after the alignment procedure, the R-peak amplitude recorded by the textile system more
closely corresponds to that of the BTL system. Additionally, the ratio of the R-peak to the T-wave in the aligned
textile signal more closely resembles the \gls{ecg} of the reference system than the unprocessed textile recording.
Nevertheless, a notable limitation persists: the T-wave remains inverted in the textile recording compared to the
reference system, showing that it is no trivial task to approximate one recording to another recording.
\subsection{Estimating Dirichlet distributions}
\begin{figure}[ht!]
\centering
  \includegraphics{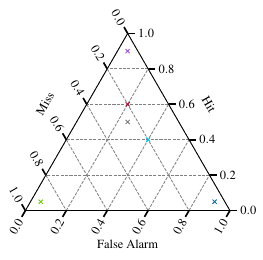}
  \caption{Triangle plot to visualize variables using a ternary plot. All points within the triangle, in our sum of all
hit, miss and false alarms sums up to 1. Purple: (Hit 0.9, Miss 0.05 and False Alarm 0.05), Green: (Hit 0.05, Miss 0.9
and False Alarm 0.05), Dark-Blue: (Hit 0.05, Miss 0.05 and False Alarm 0.9), Light-Blue: (Hit 0.4, Miss 0.2 and False
Alarm 0.4), Red: (Hit 0.6, Miss 0.1 and False Alarm 0.2), Grey: (Hit 0.5, Miss 0.25 and False Alarm 0.25)}
  \label{afig:ternary_sample}
\end{figure}
\begin{figure}[ht!]
    \centering
    \begin{subfigure}[t]{0.49\linewidth}
        \includegraphics[width=\textwidth]{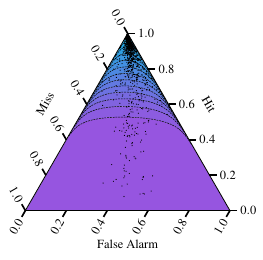}
        \caption{\centering \small BTL II vs. BTL I, $\alpha = (7.82, 0.49, 0.59)$}
        \label{sfig:btl-ecg-ii-btl-ecg-i-dirichlet}
    \end{subfigure}
    \begin{subfigure}[t]{0.49\linewidth}
        \includegraphics[width=\textwidth]{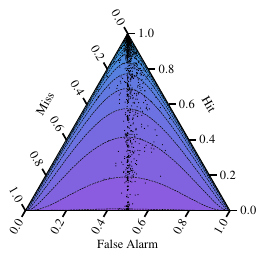}
        \caption{\centering \small BTL II vs. BTL V1, $\alpha = (1.2, 0.34, 0.36)$ }
        \label{sfig:btl-ecg-ii-btl-ecg-v1-dirichlet}
    \end{subfigure}
\caption{Log density of the Dirichlet distribution visualization and $\mathbf{\alpha}$ parameter vector of miss, hit
and false alarm values with BTL lead~II as reference and BTL leads I and V1 as evaluation.}
\label{app:fig:dirichlet_visualization}
\end{figure}

Hit Rate, Miss Rate and False Alarm Rate are mutually exclusive and sum up to one (It is either hit, miss or false
alarm). Using observations from multiple subjects and detectors, we are estimating a Dirichlet distribution given by
Equation \eqref{equ:D}:
\begin{equation}
    \mathcal{D}(\mathbf{\alpha}) = \frac{1}{\mathrm{B}(\boldsymbol\alpha)} \prod_{i=1}^K x_i^{\alpha_i - 1} \\
    \label{equ:D}
\end{equation}
where the beta function $\mathrm{B}(\boldsymbol\alpha)$ shown in Equation~\ref{equ:B} express by gamma functions
$\Gamma(\alpha)$
\begin{equation}
    \mathrm{B}(\boldsymbol\alpha) = \frac{\prod_{i=1}^K \Gamma(\alpha_i)}{\Gamma\bigl(\alpha_0\bigr)} \\
    \label{equ:B}
\end{equation}
and parameter $\alpha_0$ given by Equation~\ref{equ:A}
\begin{equation}
    \alpha_0 = \sum_{i=1}^K\alpha_i
    \label{equ:A}
\end{equation}
using maximum likelihood estimations \cite{minkaEstimatingDirichletDistribution2012}. We report the parameter vector
$\boldsymbol{\alpha} = \left(\alpha_1, \alpha_2, \alpha_3\right)$ and the visualization of the estimated Dirichlet
distribution in Sup.~Fig.~\ref{app:fig:dirichlet_visualization}.

To compare two Dirichlet distributions, we utilize the log-likelihood ratio as the test statistic. Given two
distributions  corresponding to reference device / evaluation device pairs, denoted
$\mathcal{D}_{\mathrm{Evaluation_1}}^{\mathrm{Reference}}$ and
$\mathcal{D}_{\mathrm{Evaluation_2}}^{\mathrm{Reference}}$, we first join the data underlying these distributions to
construct a joint distribution,
$\mathcal{D}_{Joint}^{\mathrm{Reference}} = (\mathcal{D}_{\mathrm{Evaluation_1}}^{\mathrm{Reference}},
\mathcal{D}_{\mathrm{Evaluation_2}}^{\mathrm{Reference}})$. Employing maximum likelihood estimation, we independently
fit Dirichlet parameters to each individual device/evaluation pair and the joint, resulting in parameter vectors
$\mathbf{\alpha}^1$ for $\mathcal{D}_{\mathrm{Evaluation_1}}^{\mathrm{Reference}}$ and $\mathbf{\alpha}^2$ for
$\mathcal{D}_{\mathrm{Evaluation_2}}^{\mathrm{Reference}}$. Additionally, we fit a parameter vector
$\mathbf{\alpha}^{Joint}$ to the combined dataset.
Finally, the log-likelihoods of each distribution, evaluated under their respective fitted parameters, are then
computed. To quantify the difference between the two distributions, we calculate the following test statistic:
\begin{equation}
\begin{split}
D = 2 \big[ & \log L(\mathcal{D}_{\mathrm{Evaluation_1}}^{\mathrm{Reference}}, \mathbf{\alpha}^1) \\
    & + \log L(\mathcal{D}_{\mathrm{Evaluation_2}}^{\mathrm{Reference}}, \mathbf{\alpha}^2) \\
    & - \log L(\mathcal{D}_{\mathrm{Joint}}^{\mathrm{Reference}}, \mathbf{\alpha}^{\mathrm{Joint}}) \big]
\end{split}
\label{equ:alpha}
\end{equation}
where $\log L$ represents the log-likelihood function. This statistic assesses the extent to which modeling the groups
separately provides a better fit, compared to modeling the reference/evaluation device and the reference/reference
device together using a single Dirichlet distribution. Parameter estimation is carried out using maximum likelihood
methods to ensure both convergence and robustness.
This likelihood ratio test for independence thus evaluates whether the two Dirichlet-distributed evaluation/reference
groups are likely to originate from the same underlying distribution or from distinct distributions.
A value of $D$ close to zero indicates that the two-distribution model does not substantially improve the fit over
the single-distribution model, suggesting that a single Dirichlet distribution adequately describes the data. In such
cases, there is insufficient evidence to justify the use of two separate distributions.

We present these likelihood ratios in Table~\ref{tab:log_ratio}, indicating that especially the distributions 
$\mathcal{D}_{\mathrm{Textile~II}}^{\mathrm{BTL~II}}$ and $\mathcal{D}_{\mathrm{Faros~II}}^{\mathrm{BTL~II}}$ can
be expressed as one distribution and are therefore closely related.
\begin{table}[!ht]
    \renewcommand{\arraystretch}{1.5}
    \caption{The likelihood ratios between two Dirichlet distributions of R-peak detection performance.}
    \label{tab:log_ratio}
    \centering
    \begin{NiceTabular}{l|cccc}
    \toprule
        \multirow{2}{*}{Phase} & $\mathcal{D}_{\mathrm{Textile~II}}^{\mathrm{BTL~II}}$& $\mathcal{D}_{\mathrm{Textile~I}}^{\mathrm{BTL~II}}$& $\mathcal{D}_{\mathrm{BTL~I}}^{\mathrm{BTL~II}}$& $\mathcal{D}_{\mathrm{BTL~I}}^{\mathrm{Faros~II}}$\\ 
        ~ & $\mathcal{D}_{\mathrm{Faros~II}}^{\mathrm{BTL~II}}$& $\mathcal{D}_{\mathrm{BTL~I}}^{\mathrm{BTL~II}}$& $\mathcal{D}_{\mathrm{BTL~V1}}^{\mathrm{BTL~II}}$& $\mathcal{D}_{\mathrm{BTL~V1}}^{\mathrm{Faros~II}}$\\
    \midrule
        Sitting Relaxed & $0.03$ & $0.90$ & $2.48$ & $2.54$ \\ 
        Sitting Video & $0.19$ & $3.84$ & $8.52$ & $7.49$ \\ 
        Sitting $n$-back & $0.04$ & $0.49$ & $6.36$ & $5.58$ \\ 
        Lying Back & $0.12$ & $1.64$ & $9.73$ & $10.51$ \\ 
        Lying Side & $0.21$ & $1.96$ & $1.08$ & $1.61$ \\ 
        Lying Side & $1.19$ & $8.87$ & $5.96$ & $5.08$ \\ 
        Lying Stomach & $0.27$ & $4.10$ & $10.02$ & $10.96$ \\ 
        Standing & $0.26$ & $5.40$ & $6.49$ & $5.97$ \\ 
        Ergometer & $0.15$ & $0.93$ & $0.25$ & $0.23$ \\ 
        Running & $1.72$ & $3.19$ & $0.02$ & $0.00$ \\ 
        Walking & $0.03$ & $0.90$ & $5.52$ & $4.98$ \\ 
    \bottomrule
    \end{NiceTabular}
\end{table}
\subsection{Training}
%Specifically, the inner folds (comprising 4 out of 5 outer folds)
%were allocated for training (four inner folds) and validation (one inner fold), while the remaining outer fold was
%reserved for testing.
To supplement our main article, we employed hyperparameter optimization using the TPESampler
\cite{watanabeTreeStructuredParzenEstimator2023}, tuning the learning rate within the range of
$3\times 10^{-4}$ to $5\times 10^{-1}$, a tree depth from 3 to 12, a number of estimators from 750 to 4000, and a gamma
between 0 and 1.5. The model with the overall best hyperparameter of the inner folds was evaluated then evaluated on
the outer test set.

\bibliographystyle{IEEEtran}
\bibliography{bibliography}

\begin{figure}[ht!]
    \centering
    % easily resize however you like!
    \resizebox{3.5in}{!}{
    
    % For organization, I like to keep the actual card in a separate file:
    \sffamily
    {
    \fbox{%
    \begin{tabular}{@{}p{\NFwidth}@{}}
    
    % Top Matter
    \multicolumn{1}{c}{\huge\bfseries Textile/Holter Dataset}\\
    \NFrule
    \NFtext{\textbf{Dataset} Signal Quality Evaulation}\\
    \NFtext{\textbf{Year of Creation} 2023}\\
    \NFtext{\textbf{Instances Per Dataset} 0 Public, 30 Private}\\
    \NFtext{\textbf{DOI} https://doi.org/XXXXX}\\
    
    % Motivation Section
    \NFRULE
    \NFline{Motivation}\\
    \NFrule
    \NFelement{Original Authors}{}{Maximilian P. Oppelt et al.}\\
    \NFelement{Original Use Case}{}{Device Quality Evaluation}\\
    \NFelement{Original Funding}{}{Digital Health and Analytics, Fraunhofer IIS}\\
    %\NFelement{Additional Field}{}{Info}\\
    
    % Composition Section
    \NFRule
    \NFline{Composition}\\
    \NFrule
    \NFelement{Sample or Complete}{}{Private dataset}\\ % Additional subjects are currently recorded that also have \gls{eeg} and movement chiar sensors.}\\
    \NFelement{Missing Data}{}{Some records with noise}\\ % Mostly complete. Some lead falloffs, missing video and missing pupil diameter due to blinking}\\
    \NFelement{Sensitive Information }{Anonymized}{, no image/video data available}\\ % Camera images and pupil images are not released to the public for privacy concerns.}\\
    %\NFelement{Additional Field}{}{Info}\\
    
    % Collection Process Section
    \NFRule
    \NFline{Collection}\\
    \NFrule
    %\NFelement{Sampling Strategy}{}{Convenience}\\
    \NFelement{Ethical Review}{}{65\_21 B on 16.03.2021}\\
    \NFelement{Ethics Committee}{}{Friedrich-Alexander-University Erlangen-Nuremberg}\\
    %\NFelement{Additional Field}{}{Info}\\
    
    % Cleaning / Labeling Section
    \NFRule
    \NFline{Cleaning and Labeling}\\
    \NFrule
    \NFelement{Cleaning Done}{}{2025-01-01}\\
    \NFelement{Labeling Done}{}{2025-01-01}\\
    %\NFelement{Additional Field}{}{Info}\\

    % Uses / Distribution Section
    \NFRule
    \NFline{Uses and Distribution}\\
    \NFrule
    \NFelement{Notable Uses}{}{Signal quality evaluation, difference in sex, projection, morphology}\\
    \NFelement{Other Uses}{}{Classification of activity, psychophysiological cognitive load}\\
%Evaluation of \gls{rppg}, signal quality tests, %
%out-of-distribution test set for e.g. emotion recognition,
%universal physiological feature extractor pretraining.}\\
    %\NFelement{Original Distribution}{}{Info}\\
    %\NFelement{Replicate Distribution}{}{Info}\\
    %\NFelement{Additional Field}{}{Info}\\
    
    % Maintenance and Evolution Section
    \NFRule
    \NFline{Maintenance and Evolution}\\
    \NFrule
    \NFelement{Corrections or Erratum}{}{None}\\
    \NFelement{Methods to Extend}{}{None}\\
    \NFelement{Replicate Maintainers}{}{Maximilian P. Oppelt}\\
    %\NFelement{Additional Field}{}{Info}\\
    
    \NFRULE
    \NFline{Breakdown}\\
    \NFrule
    \NFelement{Physiological Measurements}{}{\gls{ecg} lead I, II, V1 BTL, lead II Faros and I, II from wearable garment}\\
    \NFrule
    \NFelement{Demographics}{}{Age, sex} \\
    \NFrule
    \NFelement{Body measurements}{}{Height, weight, upper body width, shoulder width, relative electrode positioning,
upper body hairiness} \\
    \NFrule
    \NFelement{Ergonomics}{}{Quality of fit} \\
  \end{tabular}}
    
    \par} % to end centering    
  }
        \caption{A data card styled after nutrition labels. Description of wearable textile \gls{ecg} garment.}
    \label{fig:data_card}
\end{figure}

\newpage